\documentclass[11pt, a4paper]{lumia}

\usepackage[sort&compress]{natbib}
\usepackage{xspace}

\theoremstyle{plain}

\newtheorem*{proposition*}{Proposition}

\theoremstyle{definition}

\theoremstyle{definition}

\def\eqref#1{equation~\ref{#1}}

\usepackage{graphicx}           
\usepackage{tikz}               
\usepackage[edges]{forest}      

\usepackage{url}                
\usepackage{xurl}               

\usepackage{array}              
\usepackage{longtable}          
\usepackage{multirow}           
\usepackage{makecell}           
\usepackage{ragged2e}           

\usepackage{mathtools}          
\usepackage{nicefrac}           

\usepackage{algorithm}          
\usepackage{algorithmicx}       
\usepackage{algpseudocode}      
\usepackage{listings}           

\usepackage{subcaption}         
\usepackage{wrapfig}            
\usepackage[export]{adjustbox}  

\usepackage[commandnameprefix=always]{changes} 
\usepackage{xspace}             
\usepackage[normalem]{ulem}     
\usepackage{CJKutf8}            

\usepackage[tikz]{bclogo}       
\usepackage[framemethod=tikz]{mdframed} 

\usepackage{lipsum}             
\usepackage{tocloft}            
\usepackage{afterpage}          
\usepackage{bbding}             
\usepackage{epigraph}           
\usepackage{minitoc}            
\usepackage{multicol}           
\usepackage{textgreek}          
\usepackage{rotating}

\usepackage{fontawesome5}

\newcolumntype{P}[1]{>{\RaggedRight\arraybackslash}p{#1}}

\definecolor{uclablue}{RGB}{39, 116, 174}
\definecolor{bigaired}{RGB}{156, 0, 0}
\definecolor{myblue}{HTML}{598BE7}
\definecolor{mildblue}{RGB}{31,119,180}
\definecolor{sectionblue}{RGB}{70, 130, 180}
\definecolor{methodblue}{RGB}{0, 150, 136}
\definecolor{bgblue}{RGB}{245,243,253}
\definecolor{ttblue}{RGB}{91,194,224}
\definecolor{mygreen}{rgb}{0.64, 0.56, 0.88}
\definecolor{myyellow}{rgb}{0.68, 0.6, 0.1}
\definecolor{fancygreen}{rgb}{0.33, 0.68, 0.20}
\definecolor{salmon}{rgb}{0.94, 0.52, 0.49}
\definecolor{tablegreen}{rgb}{0.82, 0.94, 0.75}
\definecolor{tableblue}{rgb}{0.81, 0.90, 0.94}
\definecolor{tablered}{rgb}{0.97, 0.85, 0.85}
\definecolor{tableorange}{rgb}{0.96, 0.85, 0.81}
\definecolor{myorange}{rgb}{1.0, 0.49, 0.0}
\definecolor{tlgreen}{rgb}{0.33, 0.68, 0.20}
\definecolor{darkgreen}{RGB}{0,100,0}
\definecolor{darkred}{RGB}{200, 0, 0}
\definecolor{customyellow}{HTML}{FFFACD}
\definecolor{refinegreen}{RGB}{0, 128, 75}
\definecolor{scoregreen}{RGB}{34, 139, 34}
\definecolor{hidden-blue}{RGB}{194,232,247}
\definecolor{hidden-black}{RGB}{20,68,106}
\definecolor{yes}{HTML}{C6EFCE}
\definecolor{no}{HTML}{FFC7CE}
\definecolor{partial}{HTML}{FFEB9C}
\definecolor{external}{HTML}{D9E1F2}
\definecolor{hdr}{HTML}{F2F2F2}
\definecolor{GRPOrow}{gray}{0.96}
\definecolor{FlowRLrow}{RGB}{225,236,255}
\definecolor{FlowBlue}{RGB}{80,120,210}
\definecolor{GRPOGray}{gray}{0.35}

\hypersetup{
    colorlinks=true, 
    citecolor=uclablue, 
    linkcolor=bigaired,
    urlcolor=darkblue
}

\setlist[itemize]{leftmargin=20pt, noitemsep, topsep=0pt}

\NewDocumentCommand{\kaiyan}{mO{}}{\textcolor{purple}{\textsuperscript{\textit{kaiyan}}\textsf{\textbf{\small[#1]}}}}
\NewDocumentCommand{\yuxin}{mO{}}{\textcolor{cyan}{\textsuperscript{\textit{yuxin}}\textsf{\textbf{\small[#1]}}}}
\NewDocumentCommand{\bx}{mO{}}{\textcolor{green}{\textsuperscript{\textit{bx}}\textsf{\textbf{\small[#1]}}}}
\NewDocumentCommand{\at}{mO{}}{\textcolor{red}{\textsuperscript{\textit{AT}}\textsf{\textbf{\small[#1]}}}}
\NewDocumentCommand{\re}{mO{}}{\textcolor{blue}{\textsuperscript{\textit{RE}}\textsf{\textbf{\small[#1]}}}}
\NewDocumentCommand{\ybsun}{mO{}}{\textcolor{magenta}{\textsuperscript{\textit{youbang}}\textsf{\textbf{\small[#1]}}}}
\NewDocumentCommand{\runze}{mO{}}{\textcolor{orange}{\textsuperscript{\textit{runze}}\textsf{\textbf{\small[#1]}}}}
\NewDocumentCommand{\add}{mO{}}{\textcolor{darkgreen}{\textsuperscript{\textit{Maybe Consider Discuss}}\textsf{\textbf{[#1]}}}}

\newcommand{\cmark}{\textcolor{darkgreen}{\boldmath$\checkmark$}}
\newcommand{\xmark}{\textcolor{darkred}{\boldmath$\times$}}

\newenvironment{itemize*}%
 {\leftmargini=10pt\begin{itemize}%
  \setlength{\itemsep}{0pt}%
  \setlength{\parskip}{0pt}%
  }%
 {\end{itemize}}

\newenvironment{enumerate*}%
 {\begin{enumerate}%
  \setlength{\itemsep}{0pt}%
  \setlength{\parskip}{0pt}}%
 {\end{enumerate}}

\newcommand{\cellstatus}[1]{%
  \begingroup
  \StrTrim{#1}[\statusval]%
  \IfStrEq{\statusval}{Yes}{\cellcolor{yes}\cmark}{}%
  \IfStrEq{\statusval}{No}{\cellcolor{no}\xmark}{}%
  \IfBeginWith{\statusval}{Yes (}{\cellcolor{yes}\cmark~\textit{\statusval\unskip}}{}%
  \IfStrEq{\statusval}{Partial}{\cellcolor{partial}\textbf{Partial}}{}%
  \IfStrEq{\statusval}{External}{\cellcolor{external}\textbf{External}}{}%
  \endgroup
}

\newtcolorbox{myboxi}[1][]{
  breakable,
  title=#1,
  colback=red!5,
  colbacktitle=red!5,
  coltitle=black,
  fonttitle=\bfseries,
  bottomrule=0pt,
  toprule=0pt,
  leftrule=2pt,
  rightrule=2pt,
  titlerule=0pt,
  arc=0pt,
  outer arc=0pt,
  colframe=red,
}

\newtcolorbox{myboxnote}[1][]{
  breakable,
  title=#1,
  colback=orange!0,
  colbacktitle=orange!0,
  coltitle=black,
  fonttitle=\bfseries,
  bottomrule=0pt,
  toprule=0pt,
  leftrule=2pt,
  rightrule=2pt,
  titlerule=0pt,
  arc=0pt,
  outer arc=0pt,
  colframe=orange,
}

\newtcolorbox{myboxii}[1][]{
  breakable,
  freelance,
  title=#1,
  colback=white,
  colbacktitle=white,
  coltitle=black,
  fonttitle=\bfseries,
  bottomrule=0pt,
  boxrule=0pt,
  colframe=white,
  overlay unbroken and first={
  \draw[red!75!black,line width=3pt]
    ([xshift=5pt]frame.north west) -- 
    (frame.north west) -- 
    (frame.south west);
  \draw[red!75!black,line width=3pt]
    ([xshift=-5pt]frame.north east) -- 
    (frame.north east) -- 
    (frame.south east);
  },
  overlay unbroken app={
  \draw[red!75!black,line width=3pt,line cap=rect]
    (frame.south west) -- 
    ([xshift=5pt]frame.south west);
  \draw[red!75!black,line width=3pt,line cap=rect]
    (frame.south east) -- 
    ([xshift=-5pt]frame.south east);
  },
  overlay middle and last={
  \draw[red!75!black,line width=3pt]
    (frame.north west) -- 
    (frame.south west);
  \draw[red!75!black,line width=3pt]
    (frame.north east) -- 
    (frame.south east);
  },
  overlay last app={
  \draw[red!75!black,line width=3pt,line cap=rect]
    (frame.south west) --
    ([xshift=5pt]frame.south west);
  \draw[red!75!black,line width=3pt,line cap=rect]
    (frame.south east) --
    ([xshift=-5pt]frame.south east);
  },
}

\tcbset{
  takeawaysbox/.style={
    title=Takeaways,
    colback=lightblue!80,
    colframe=black,
    fonttitle=\bfseries\small,
    coltitle=white,
    colbacktitle=black,
    enhanced,
    attach boxed title to top left={xshift=2.5mm,yshift=-2.5mm},
    boxed title style={rounded corners, size=small, colframe=black, colback=black},
    width=\linewidth,
    arc=3.5mm
  }
}

\mdfdefinestyle{mystyle}{%
  rightline=true,
  innerleftmargin=10,
  innerrightmargin=10,
  outerlinewidth=3pt,
  topline=false,
  rightline=true,
  bottomline=false,
  skipabove=\topsep,
  skipbelow=\topsep
}

\tikzset{%
    every node/.style={font=\tiny},
    parent/.style =          {align=center,text width=2cm,rounded corners=3pt, line width=0.3mm, fill=gray!10,draw=gray!80},
    child/.style =           {align=center,text width=2.0cm,rounded corners=3pt, fill=blue!10,draw=blue!80,line width=0.3mm},
    grandchild/.style =      {align=center,text width=2cm,rounded corners=3pt},
    greatgrandchild/.style = {align=center,text width=1.5cm,rounded corners=3pt},
    greatgrandchild2/.style = {align=center,text width=1.5cm,rounded corners=3pt},    
    referenceblock/.style =  {align=center,text width=1.5cm,rounded corners=2pt},
    pretrain/.style =           {align=center,text width=2.0cm,rounded corners=3pt, fill=blue!10,draw=blue!80,line width=0.3mm},   
    pretrain_work/.style =           {align=center, text width=8.5cm,rounded corners=3pt, fill=blue!10,draw=blue!0,line width=0.3mm},  
    template/.style =           {align=center,text width=2.0cm,rounded corners=3pt, fill=red!10,draw=red!80,line width=0.3mm},   
    template_work/.style =           {align=center,text width=8.5cm,rounded corners=3pt, fill=red!10,draw=red!0,line width=0.3mm},    
    answer/.style =           {align=center,text width=2.0cm,rounded corners=3pt, fill= cyan!10,draw= cyan!80,line width=0.3mm},   
    answer_work/.style =           {align=center,text width=8.5cm,rounded corners=3pt, fill= cyan!10,draw= cyan!0,line width=0.3mm},      
    multiple/.style =           {align=center,text width=2.0cm,rounded corners=3pt, fill= orange!10,draw= orange!80,line width=0.3mm},   
    multiple_work/.style =           {align=center,text width=8.5cm,rounded corners=3pt, fill= orange!10,draw= orange!0,line width=0.3mm},        
    tuning/.style =           {align=center,text width=2.0cm,rounded corners=3pt, fill= magenta!10,draw= magenta!80,line width=0.3mm},   
    tuning_work/.style =           {align=center,text width=8.5cm,rounded corners=3pt, fill= magenta!10,draw= magenta!0,line width=0.3mm},          
}

\newcommand{\lstbg}[3][0pt]{{\fboxsep#1\colorbox{#2}{\strut #3}}}

\lstdefinelanguage{diff}{
  basicstyle=\ttfamily\small,
  morecomment=[f][\lstbg{red!20}]-,
  morecomment=[f][\lstbg{green!20}]+,
}

\lstdefinelanguage{diffpython}{
  language=diff,
  morekeywords={def, if, else, for, while, return, import, from, as, class, with, try, except, finally, raise, lambda, and, or, not, in, is, None, True, False},
  morecomment=[l]{\#},
  morestring=[b]",
  morestring=[b]',
}

\setheadertext{NCP-ArchPreview Technical Report}

\title{NCP-ArchPreview Technical Report: \\ Moving towards Latent Space Language Models through Next Concept Prediction}
\setheadertitle{NCP-ArchPreview Technical Report}

\author{%
The Intern-NCP Team \\
\Authfont Shanghai AI Lab; LUMIA Lab, Shanghai Jiao Tong University

\begin{minipage}{\linewidth}
\raggedright
\normalfont\small
\setlength{\parskip}{0pt}
\noindent\strut
\makebox[1.8em][c]{%
\raisebox{-0.28em}[0pt][0pt]{%
\includegraphics[height=0.9em]{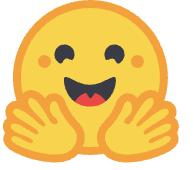}}%
}
\hspace{0.35em}%
\href{https://huggingface.co/collections/ArchSpace-Collection/ncp-archpreview}
{\nolinkurl{https://huggingface.co/collections/ArchSpace-Collection/ncp-archpreview}}
\par\vspace{0.4em}
\noindent\strut
\makebox[1.8em][c]{%
\raisebox{-0.28em}[0pt][0pt]{%
\includegraphics[height=1.25em]{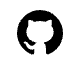}}%
}
\hspace{0.35em}%
\href{https://github.com/InternLM/lmdeploy}
{\nolinkurl{https://github.com/InternLM/lmdeploy}}
\par\vspace{0.4em}
\noindent\strut
\makebox[1.8em][c]{%
\raisebox{-0.28em}[0pt][0pt]{%
\includegraphics[height=1.25em]{assets/github-icon.png}}%
}
\hspace{0.35em}%
\href{https://github.com/LUMIA-Group/ncp_olmo_eval}
{\nolinkurl{https://github.com/LUMIA-Group/ncp_olmo_eval}}
\par
\end{minipage}

}

\begin{document}


\begin{abstract}

We introduce \textbf{\texttt{NCP-ArchPreview}}, a latent-space language model that pushes autoregressive pretraining beyond standard next-token prediction (NTP). Alongside NTP, the model learns through Next Concept Prediction (NCP) to predict discrete concepts that span multiple tokens, introducing an explicit and more challenging concept-level objective while preserving standard token-level autoregressive generation. \texttt{NCP-ArchPreview} builds a latent space by constructing a product-quantized concept vocabulary directly from its hidden states, and subsequently learns to predict future concepts via a dedicated Concept Module. These predicted concepts are then fed back to the token level to guide subsequent generation, with NTP and NCP trained jointly end-to-end. We scale this architecture to 8.9B parameters and train it on 5.73T tokens from the \texttt{Dolma-3} dataset, marking the largest demonstration of a latent-space language model to date. Remarkably, by consuming only 51.3\% of the total training tokens, \texttt{NCP-ArchPreview} achieves the final pretraining loss of \texttt{OLMo-3-7B}. Following full pretraining, it outperforms \texttt{OLMo-3-7B} by 2.45 points on the downstream macro-average, including a notable 5.99-point gain on GSM8K. Controlled experiments isolate a clear progression of performance gains stemming from both the latent architecture and the NCP objective. Furthermore, utilizing only 85\% of the standard computation, \texttt{NCP-ArchPreview} approaches the training loss of a strictly parameter-aligned 8.9B baseline.
The learned latent space remains highly valuable after the pretraining stage: updating just the 17M-parameter VQ module yields a novel, lightweight interface for domain adaptation, while a simple injection of concept representations into a DFlash2 drafter improves the mean accepted length by 4.17\% with negligible overhead. 
By proving the viability of joint token and concept modeling at scale, these results position latent-space prediction not merely as an auxiliary objective, but a highly efficient and scalable architectural blueprint for next-generation foundation models.
We release the Stage-1 checkpoints obtained every 100,000 training steps, the corresponding drafter models, and the final Stage-1 / Stage-2 checkpoint.

\end{abstract}

\maketitle



\section{Introduction}
\label{sec:introduction}

\begin{figure}[tbp]
    \centering
    \begin{subfigure}{\textwidth}
        \centering
        \includegraphics[width=0.88\linewidth]{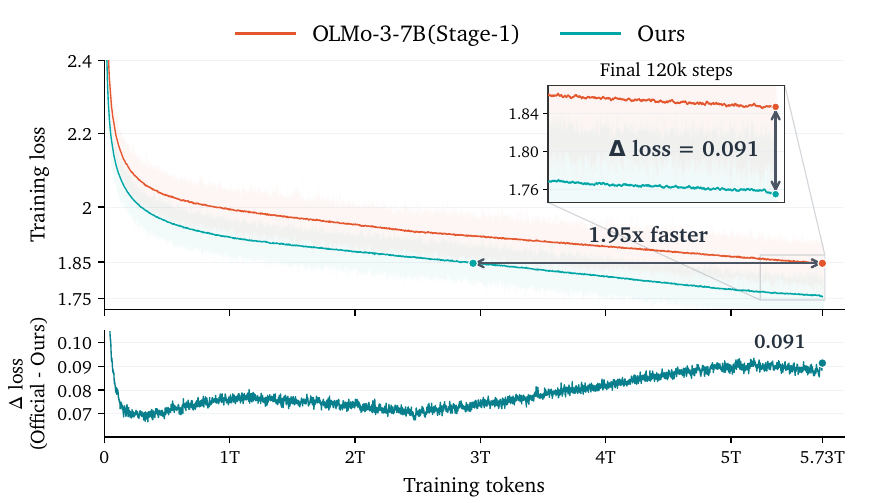}
        \caption{Stage-1 loss curves for \texttt{OLMo-3-7B} and \texttt{NCP-ArchPreview} across all 5.73T pretraining tokens. Despite trained on exactly the same data, \texttt{NCP-ArchPreview} converges 1.95× faster and achieves a 0.091 lower loss over the final steps.}
        \label{fig:stage1-loss}
    \end{subfigure}
    \hfill
    \begin{subfigure}{\textwidth}
        \centering
        \includegraphics[width=0.88\linewidth]{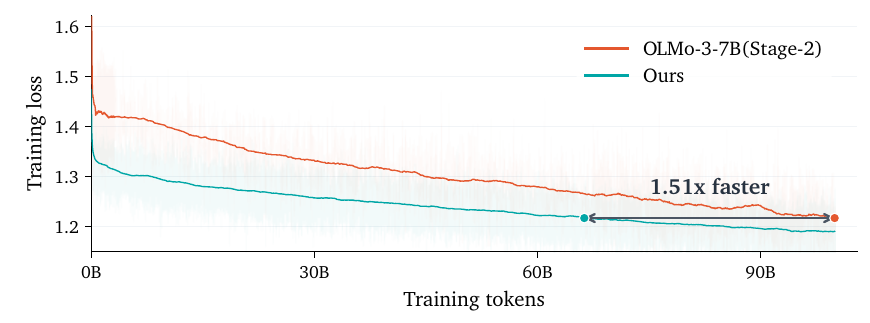}
        \caption{Stage-2 loss curves for \texttt{OLMo-3-7B} and \texttt{NCP-ArchPreview} on the same data, showing that \texttt{NCP-ArchPreview} converges 1.51× faster with less fluctuations while maintaining a lower training loss.}
        \label{fig:stage2-loss}
    \end{subfigure}
    \caption{Comparison of training dynamics between \texttt{OLMo-3-7B} and our \texttt{NCP-ArchPreview} during pretraining Stage-1 and Stage-2.}
    \label{fig:stage-loss}
\end{figure}

Recent progress in generative AI has demonstrated that the representation space in which a model learns can matter as much as its parameter scale. In visual synthesis, latent diffusion shifts generation from raw pixels into compact continuous representations, dramatically boosting modeling efficiency and scalability~\citep{rombach2022highresolution,blattmann2023align}. Modern language models similarly induce high-level abstractions, such as semantic concepts and latent world representations, in their hidden states~\citep{li2023emergentworldrepresentations,gurnee2024language,park2024linearrepresentation}. Under standard Next Token Prediction (NTP), however, these abstractions arise purely as an indirect byproduct: supervision is strictly confined to granular tokens, lacking explicit objectives that guide how semantic structure unfolds across multi-token spans. To bridge this gap, we introduce \textbf{\texttt{NCP-ArchPreview}}, a novel latent-space foundation model architecture that incorporates the direct prediction of discrete concepts spanning over multiple tokens into the pretraining stage. By jointly pretraining with both NTP and Next Concept Prediction (NCP), \texttt{NCP-ArchPreview} scales latent-space language modeling up to 5.73T pretraining tokens, establishing the architectural viability and scalability of this paradigm at frontier scale.

Prior architectural efforts can be distinguished along two dimensions: how latent representations are formed, and what they are trained to represent. In terms of latent representation, the Hourglass Transformer~\citep{hourglass}, ContextLM~\citep{dai2025contextlevellanguagemodelinglearning}, and MegaByte~\citep{yu2023megabytepredictingmillionbytesequences} use fixed hierarchical structures, with predefined multiscale resolutions or byte patches. In contrast, the Byte Latent Transformer (BLT)~\citep{pagnoni2024bytelatenttransformerpatches}, DLCM~\citep{qu2026dynamiclargeconceptmodels}, and H-Net~\citep{hwang2025dynamicchunkingendtoendhierarchical} use dynamic or input-adaptive chunking, allowing latent boundaries or patch sizes to vary with the input. A separate set of methods introduces new latent prediction targets or training objectives. Large Concept Model~\citep{lcmteam2024largeconceptmodelslanguage} predicts representations in a shared continuous sentence-embedding space; and ConceptLM~\citep{liu2026conceptpredictiondiscretelatent} predicts discrete concepts from a learned concept vocabulary.

Joint-Embedding Predictive Architectures (JEPA)~\citep{lecun2022path} offer a principled alternative: predict future representations directly in latent space to capture invariant, semantic structures. This paradigm has driven breakthroughs in image representation learning (I-JEPA)~\citep{assran2023selfsupervisedlearningimagesjointembedding} and large-scale video modeling (V-JEPA and V-JEPA 2)~\citep{bardes2024vjepa,assran2025vjepa2}. In language modeling, however, learning explicitly upon latent targets remains underexplored. While Multi-Token Prediction (MTP)~\citep{gloeckle2024betterfasterlarge} supervises multiple future positions, its loss is still tethered to individual surface tokens. \texttt{NCP-ArchPreview} extends this principle to language modeling via Next Concept Prediction (NCP)~\citep{liu2026conceptpredictiondiscretelatent}, which predicts latent representations directly over a learned concept vocabulary.

Built upon the \texttt{OLMo-3-7B} backbone~\citep{olmo2026olmo3}, \texttt{NCP-ArchPreview} establishes an 8.9B-parameter realization of this architecture. It inserts an 8-layer Concept Module between a 16-layer Token Encoder and a 16-layer Token Decoder, adopts product quantization to construct an expressive discrete concept space, and incorporates multiway dynamic dense connections adapted from MUDDFormer~\citep{xiao2025muddformer} to facilitate hierarchical residual routing. Extensive pretraining across 5.73T tokens reveals substantial optimization efficiency: \texttt{NCP-ArchPreview} matches the final pretraining loss of \texttt{OLMo-3-7B} using only 51.3\% of its training tokens ($1.95\times$ convergence speedup), and outperforms \texttt{OLMo-3-7B} by 2.45 points on downstream macro-averages (+5.99 on GSM8K) after pretraining. Compute-aligned and parameter-aligned ablations confirm that these gains stem fundamentally from the latent hierarchy and the NCP objective rather than superficial capacity increases, yielding a $1.74\times$ improvement in Pareto compute efficiency. Beyond pretraining, the learned concept space serves as an efficient 17M-parameter interface for lightweight domain adaptation and accelerates inference by +4.17\% mean accepted length when integrated into a DFlash2 speculative drafter~\citep{inco2026dflash2}.

To accelerate open-source research on non-vanilla foundation architectures, with this paper we formally release \textbf{\texttt{NCP-ArchPreview}} as a fully pretrained latent-space foundation model, including its model weights, inference scripts, training recipes, and intermediate evaluation checkpoints. In the remainder of this report, we provide a comprehensive technical account of this release: Section~\ref{sec:architecture} details the modular architecture and hierarchical residual routing; Section~\ref{sec:training} introduces the joint NTP-NCP training objective; and Section~\ref{sec:evaluation} presents empirical validations across trillion-token pretraining, component ablations, scaling laws, and downstream evaluations. We further investigate practical training considerations, including attention-logit stabilization under matrix optimizers and capability-aware proxy metrics for mid-training data curation, before discussing downstream adaptation mechanisms, multi-token prediction synergies, and future directions for latent-space language modeling.


\section{Model Architecture}
\label{sec:architecture}

We introduce \texttt{NCP-ArchPreview}, a latent-space language model that couples token-level modeling with concept-level prediction in a discrete latent space. We construct a discrete concept vocabulary from intermediate token representations, use a Concept Module to predict future concepts, and feed those predictions back into the token stream. Hierarchical residual further mixes information across depths and modules. The following subsections define these components.

\subsection{Overview}
\label{subsec:architecture-overview}

Our model introduces a latent-space prediction to a standard NTP model.  The backbone is organized into three modules: a Token Encoder, a Concept Module, and a Token Decoder. The Token Encoder produces token representations, and then the model compresses contiguous groups of tokens into continuous concept representations. The Concept Module predicts the next concept from the preceding concepts, while the Token Decoder uses the predicted concepts together with token-level representations to predict the next token. In addition, our hierarchical residual connects layers within each module and across the three modules, allowing information to travel across depth and hierarchy.

Let $k$ be the concept compression factor. Let $M=\lfloor T/k \rfloor$. The Token Encoder maps the input sequence to token-level hidden states$\mathbf{h}_{1:T}$. Then we use mean pooling over each group of $k$ states to produce a continuous concept sequence $\mathbf{c}_{1:M}$. We use Vector Quantization (VQ) to learn a finite concept space, forming the concept vocabulary~\citep{Vector_Quantization}. The Concept Module then predicts a differentiable representation of the next concept by a weighted combination of codebook entries. Before being injected into the Token Decoder, the predicted concept sequence is repeated at token resolution, shifted to preserve causality. The corresponding training objectives are described in Sec.~\ref{subsec:model_training}.

\begin{figure*}[t]
    \centering
    \includegraphics[width=0.85\linewidth]{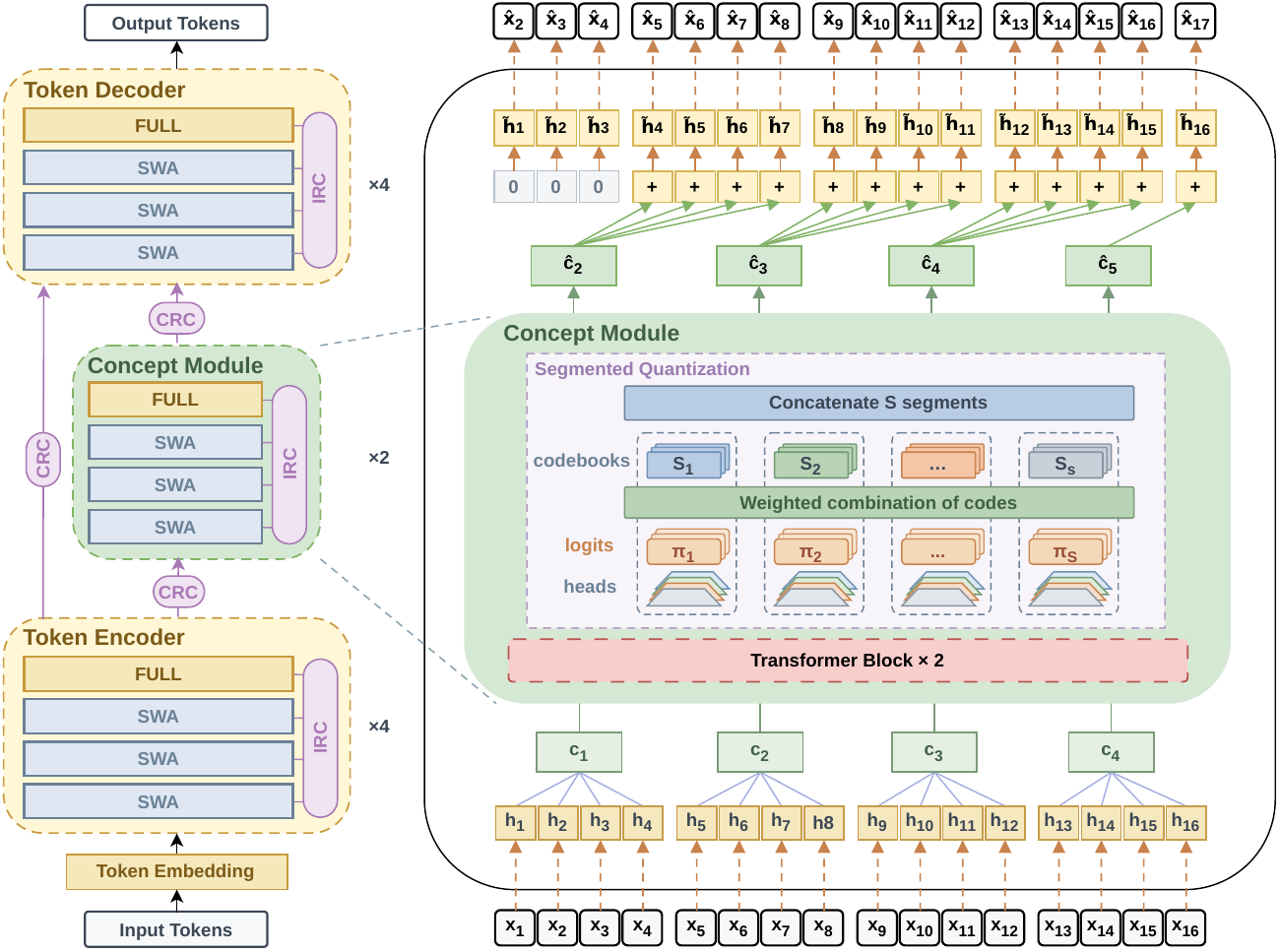}
    \caption{Overview of \texttt{NCP-ArchPreview}. The token-level module (comprising the Token Encoder and Token Decoder) maintains token-level representations, ensuring the overall input and output remain consistent with canonical autoregressive decoding. In the intermediate layers, the Concept Module compresses token spans into concepts and predicts the next concept via weighted combinations of entries from a vector-quantized concept vocabulary, which is learned from the Token Encoder's hidden states. Hierarchical residual connections, including Cross-Module Residual Connections (CRC) and Intra-Module Residual Connections (IRC), allow information to flow smoothly across layers and abstraction levels.}

    \label{fig:main_arc}
\end{figure*}

\subsection{Learning the Discrete Concept Vocabulary with Vector Quantization}
\label{subsec:concept-construction}

The VQ objective encourages the codebook entries to cover the distribution of continuous concept representations, thereby providing a structured target space for prediction. Then, the Concept Module predicts the next concept in this learned space. It produces a probability distribution over each codebook, whose weighted combination gives a differentiable concept prediction. This subsection describes the codebook learning.

The Token Encoder is the first module of the backbone and produces token-level hidden states
\begin{equation}
    \mathbf{h}_{1:T}
    = \operatorname{TokenEncoder}_{\theta_e}(x_{1:T}),
    \qquad \mathbf{h}_t \in \mathbb{R}^{d}.
    \label{eq:encoder-hidden-states}
\end{equation}
These hidden states serve three purposes: they are consumed by the Token Decoder, compressed into concept-level representations for the Concept Module, and used to learn the concept codebooks.

For the $m$-th concept, we aggregate a contiguous group of $k$ token states using mean pooling:
\begin{equation}
    \mathbf{c}_m
    = f_c\!\left(\mathbf{h}_{(m-1)k+1:mk}\right)
    = \frac{1}{k}\sum_{i=1}^{k}
      \mathbf{h}_{(m-1)k+i},
    \qquad \mathbf{c}_m \in \mathbb{R}^{d}.
    \label{eq:concept-mean-pooling}
\end{equation}
The resulting concept sequence has length $M=\lfloor T/k \rfloor$, which is shorter than the token sequence by a factor of $k$.

We use VQ to map each continuous concept representation to a finite set of representative vectors. To increase the capacity of this discrete concept space without using a large monolithic codebook, we adopt product quantization~\citep{jegou2010product}. Each concept vector is partitioned into $S$ feature segments:
\begin{equation}
    \mathbf{c}_m
    = \operatorname{concat}\!\left(
      \mathbf{c}_m^1,\ldots,\mathbf{c}_m^S\right),
    \qquad \mathbf{c}_m^s \in \mathbb{R}^{d/S}.
    \label{eq:concept-segmentation}
\end{equation}

For segment $s$, let $\mathcal{E}^s=\{\mathbf{e}_1^s,\ldots,\mathbf{e}_N^s\}$ denote its codebook. We assign each continuous segment to its nearest codeword:
\begin{align}
    n_m^s
    &= \arg\min_{n\in\{1,\ldots,N\}}
       \left\|\mathbf{c}_m^s-\mathbf{e}_n^s\right\|_2^2, \\
    \mathbf{d}_m^s
    &= \mathbf{e}_{n_m^s}^s.
    \label{eq:product-quantization}
\end{align}
The complete quantized concept is obtained by concatenating the quantized
segments:
\begin{equation}
    \mathbf{d}_m
    = \operatorname{concat}
      \left(\mathbf{d}_m^1,\ldots,\mathbf{d}_m^S\right).
\end{equation}
Although each segment codebook contains only $N$ entries, product quantization defines $N^S$ possible combinations of segment codewords. It therefore substantially increases the capacity of the discrete concept space while keeping the codebooks small. The quantization loss and its stop-gradient operations are given in Sec.~\ref{subsec:model_training}.

\subsection{Predicting the Next Concept with the Learned Vocabulary}
\label{subsec:concept-prediction}

Given the learned codebooks from Sec.~\ref{subsec:concept-construction}, the Concept Module predicts the next concept autoregressively in the resulting structured latent space. During training, it consumes the continuous concept history $\mathbf{c}_{<m}$ and is supervised toward the next concept. At inference time, previously predicted concepts are fed back autoregressively. Given the history $\mathbf{c}_{<m}$, a stack of Transformer layers produces a latent hidden state
\begin{equation}
    \mathbf{u}_m
    = \operatorname{ConceptModule}_{\theta_c}(\mathbf{c}_{<m}).
    \label{eq:concept-module-hidden}
\end{equation}

For every product quantization segment $s$, a segment-specific prediction head produces a distribution over the codebook entries $\mathcal{E}^s$:
\begin{equation}
    \boldsymbol{\pi}_m^s
    =
    \operatorname{softmax}\!\left(
      \operatorname{PredictionHead}_c^s(\mathbf{u}_m)
    \right),
    \qquad \boldsymbol{\pi}_m^s \in \mathbb{R}^{N}.
    \label{eq:concept-head}
\end{equation}

Rather than taking an argmax or sampling a codeword, we form a differentiable predicted segment by taking the expectation under this distribution:
\begin{equation}
    \hat{\mathbf{c}}_m^s
    = \sum_{n=1}^{N}\pi_{m,n}^s\mathbf{e}_n^s.
    \label{eq:predicted-concept-segment}
\end{equation}

The predicted concept is the concatenation of all segment-wise predictions:
\begin{equation}
    \hat{\mathbf{c}}_m
    = \operatorname{concat}\!\left(
      \hat{\mathbf{c}}_m^1,\ldots,\hat{\mathbf{c}}_m^S
      \right).
    \label{eq:predicted-concept}
\end{equation}

This weighted-codeword construction keeps the prediction path fully differentiable while restricting it to the linear combination of the learned codebooks. The resulting NCP objective therefore supervises prediction in a constrained latent space instead of regressing to an unconstrained, freely drifting continuous target.

\subsection{Injecting Predicted Concepts into the Token Stream}
\label{subsec:concept-injection}

The span of predicted concepts must align with the token-level states consumed by the Token Decoder.  We first repeat every predicted concept $k$ times and then apply a causal shift.  Let $\Delta=k$ denote the shift used by our implementation, for token-level prediction positions $t=1,\ldots,T-1$, the token-level concept signal is

\begin{equation}
    \mathbf{b}_t=
    \begin{cases}
        \mathbf{0}, & 1\leq t<\Delta,\\[2pt]
        \hat{\mathbf{c}}_{\left\lfloor (t-\Delta)/k\right\rfloor+2},
        & \Delta\leq t<T.
    \end{cases}
    \label{eq:causal-concept-broadcast}
\end{equation}
The shift is introduced to prevent information leakage: each predicted concept is injected only after the token states used to predict its target have been processed.

The concept and token hidden states share the same hidden dimension $d$; so we directly fuse the two embeddings, with an element-wise residual addition:
\begin{equation}
    \widetilde{\mathbf{h}}_t
    =\mathbf{h}_t+\mathbf{b}_t.
    \label{eq:token-concept-fusion}
\end{equation}
The Token Decoder then predicts the next token autoregressively from the fused states,
\begin{equation}
    p(x_{t+1}\mid x_{\leq t})
    =P_{\theta_d}\!\left(\widetilde{\mathbf{h}}_{\leq t}\right).
    \label{eq:decoder-token-probability}
\end{equation}

\subsection{Hierarchical Residual Connections}
\label{subsec:hierarchical-residual}

The Token Encoder, Concept Module, and Token Decoder operate at different depths and sequence granularities. We therefore introduce hierarchical residual connections to facilitate information flow both within and across modules. Inspired by the dynamic dense connections of MUDDFormer~\citep{xiao2025muddformer}, our design uses hidden-state-dependent connection weights while retaining a single hidden stream in each module. The proposed connections have two complementary components: Intra-Module Residual Connections (IRC), which combine representations from different layers within the same module, and
Cross-Module Residual Connections (CRC), which transfer representations between modules.

\subsubsection{Intra-Module Residual Connections}
\label{subsubsec:intra-module-residual}

Let $s\in\{\mathrm{TokenEncoder},\mathrm{ConceptModule},\mathrm{TokenDecoder}\}$ denote a module. For layer $\ell$, let $\mathbf{H}_\ell^s$ be the input to its Transformer block and let
\begin{equation}
    \mathbf{R}_\ell^s
    =
    F_\ell^s(\mathbf{H}_\ell^s)
    \label{eq:irc-block-output}
\end{equation}
denotes the block output before the residual addition. In a standard Transformer, the next hidden state is obtained by adding the block output to the current state:
\begin{equation}
    \mathbf{H}_{\ell+1}^s
    =
    \mathbf{H}_\ell^s+\mathbf{R}_\ell^s.
\end{equation}

Our Intra-Module Residual Connections (IRC) generalize this fixed depth-wise addition by allowing the next state to combine representations from multiple depths. At layer $\ell$, we construct the candidate set
\begin{equation}
    \mathcal{X}_\ell^s
    =
    \left\{
      \mathbf{H}_1^s,\,
      \mathbf{H}_1^s+\mathbf{R}_1^s,\,
      \ldots,\,
      \mathbf{H}_\ell^s+\mathbf{R}_\ell^s
    \right\}.
    \label{eq:irc-candidates}
\end{equation}
A lightweight MLP applied to the current block output $\mathbf{R}_\ell^s$ produces one coefficient for each candidate state:
\begin{equation}
    \mathbf{w}_\ell^s
    =
    \operatorname{MLP}_\ell^s(\mathbf{R}_\ell^s).
    \label{eq:irc-coefficients}
\end{equation}
The next hidden state is their weighted combination:
\begin{equation}
    \mathbf{H}_{\ell+1}^s
    =
    \sum_{j=1}^{\ell+1}
    w_{\ell,j}^s\mathbf{X}_{\ell,j}^s.
    \label{eq:irc-mixing}
\end{equation}
The coefficients are not normalized, so IRC can represent signed combinations of states from different depths. We initialize the MLP such that $\mathbf{w}_\ell^s=[0,\ldots,0,1]$. Thus, IRC initially selects the most recent residual-updated state and reduces to the standard residual
connection, while training can learn to incorporate information from earlier layers.

\subsubsection{Cross-Module Residual Connections}
\label{subsubsec:cross-module-residual}

Cross-Module Residual Connections (CRC) transfer information between modules. Let
\[
    \mathcal{S}^u=\{\mathbf{S}_j^u\}_{j=1}^{K}
\]
denote the representations exported by a source module $u$, and let $\mathbf{T}_\ell^s$ denote the state of a target module $s$. When the two modules operate at different sequence granularities, the source representations are first aligned by chunking or repetition. A lightweight MLP applied to the target state then produces normalized coefficients over the source depths:
\begin{equation}
    \boldsymbol{\alpha}_\ell^{s\leftarrow u}
    =
    \operatorname{softmax}\!\left(
      \operatorname{MLP}_{\ell}^{s\leftarrow u}
      (\mathbf{T}_\ell^s)
    \right),
    \qquad
    \mathbf{M}_\ell^{s\leftarrow u}
    =
    \sum_{j=1}^{K}
    \alpha_{\ell,j}^{s\leftarrow u}
    \operatorname{LN}(\mathbf{S}_j^u).
    \label{eq:crc-mixing}
\end{equation}

The resulting representation is added to the target stream using a learned diagonal scaling:
\begin{equation}
    \mathbf{T}_{\ell,\mathrm{out}}^s
    =
    \mathbf{T}_\ell^s
    +
    \mathbf{D}_\ell^{s\leftarrow u}
    \odot
    \mathbf{M}_\ell^{s\leftarrow u},
    \qquad
    \mathbf{D}_\ell^{s\leftarrow u}\in\mathbb{R}^{d}.
    \label{eq:crc-update}
\end{equation}

We use three cross-module connections:
\begin{gather*}
    \mathrm{TokenEncoder} \rightarrow \mathrm{ConceptModule} \\
    \mathrm{TokenEncoder} \rightarrow \mathrm{TokenDecoder} \\
    \mathrm{ConceptModule} \rightarrow \mathrm{TokenDecoder}
\end{gather*}
The ConceptModule-to-TokenDecoder connection follows the same causal shift as the predicted-concept pathway and therefore does not expose future-token information. The diagonal scalings are initialized with small values, so the model starts close to the original backbone and gradually learns to exploit
the additional cross-module connections. Together, IRC and CRC define the hierarchical residual connection mechanism.
\section{Training}
\label{sec:training}
\label{subsec:model_training}

\subsection{End-to-End Training Overview}
\label{subsec:training-overview}

The Concept Module is trained jointly with the token-level model. For an input sequence $x_{1:T}$, the Token Encoder first produces token-level states $\mathbf{h}_{1:T}$, and the pooling operator in Sec.~\ref{subsec:concept-construction} produces continuous concept states $\mathbf{c}_{1:M}$.  The VQ module assigns each concept to a structured codebook representation, the Concept Module predicts the next concept from the concept history, and the causally shifted predictions are injected into the Token Decoder. The resulting token-level and concept-level predictions are under an end-to-end optimization of the concept pathway. The codebooks, Token Encoder, Concept Module, and Token Decoder are updated together by NTP and NCP objectives.

\subsection{Learning the Discrete Concept Vocabulary}
\label{subsec:vq-objective}

Our VQ codebooks provide a structured target space for concept prediction. For each product quantization segment, the codebook is trained to fit the distribution of continuous concept representations:
\begin{equation}
    \mathcal{L}_{\mathrm{VQ}}
    =
    \frac{1}{MS}
    \sum_{m=1}^{M}
    \sum_{s=1}^{S}
    \left\|
      \operatorname{sg}(\mathbf{c}_m^s)
      -
      \mathbf{d}_m^s
    \right\|_2^2,
    \label{eq:vq-loss}
\end{equation}
where $\operatorname{sg}(\cdot)$ denotes stop-gradient and $\mathbf{d}_m^s$ is the nearest codebook entry assigned to $\mathbf{c}_m^s$. The stop-gradient operation prevents the VQ objective from updating the Token Encoder. Consequently, this objective only moves the selected codebook entries toward the continuous concept representations, allowing the codebooks to track the latent distribution without directly altering the token-level hidden states.

\subsection{Next Concept Prediction}
\label{subsec:ncp-objective}

The goal of next-concept prediction is to model dependencies between successive groups of tokens at the concept level. Because each concept summarizes multiple token positions, this objective provides supervision over dependencies spanning multiple tokens. A direct regression objective over continuous concept representations would allow the model to produce arbitrary vectors in the representation space. Such an objective specifies only the distance to the target and does not define which outputs constitute valid concept representations.

We therefore parameterize the prediction through the learned VQ codebooks. For each product quantization segment, the Concept Module predicts a probability distribution over the corresponding codebook entries. The entries are then combined according to the predicted distribution, yielding a  differentiable concept representation. This design provides a structured set of reference representations while avoiding a non-differentiable discrete selection step.

Specifically, we train the Concept Module to predict the next continuous concept using a mean-squared error objective:
\begin{equation}
    \mathcal{L}_{\mathrm{NCP}}
    =
    \frac{1}{M-1}
    \sum_{m=2}^{M}
    \left\|
      \hat{\mathbf{c}}_m-\operatorname{sg}(\mathbf{c}_m)
    \right\|_2^2.
    \label{eq:ncp-loss}
\end{equation}
Here, $\mathbf{c}_m$ is the continuous concept computed from the $m$-th token group, while $\hat{\mathbf{c}}_m$ is predicted from the preceding concept history. 

The NCP loss updates both the Concept Module and the Token Encoder. The target $\mathbf{c}_m$ is detached. The Token Encoder receives NCP gradients only through the preceding concept representations
$\mathbf{c}_{<m}$ supplied to the Concept Module. This encourages the Token Encoder to retain information that is useful for predicting the next concept.

\subsection{Next Token Prediction}
\label{subsec:ntp-objective}

The fused representation $\widetilde{\mathbf{h}}_t$ from
Eq.~\ref{eq:token-concept-fusion} is trained with the usual causal
next-token objective:
\begin{equation}
    \mathcal{L}_{\mathrm{NTP}}
    =-\frac{1}{T-1}\sum_{t=1}^{T-1}
      \log p_{\theta_d}\!\left(x_{t+1}\mid
      \widetilde{\mathbf{h}}_{\leq t}\right).
    \label{eq:ntp-loss}
\end{equation}
The zero prefix and shift ensure that the NTP loss never conditions on a concept computed from the tokens whose prediction it is meant to supervise. NTP provides dense supervision for every weight in the model. It updates the token pathway and the concept pathway while preserving the standard autoregressive language-modeling behavior of the backbone.

\subsection{Joint Objective and Gradient Flow}
\label{subsec:joint-objective}

We jointly optimize the model with
\begin{equation}
    \mathcal{L}_{\mathrm{total}}
    =
    \mathcal{L}_{\mathrm{NTP}}
    +\alpha L_{\mathrm{NCP}}
    +\beta L_{\mathrm{VQ}},
    \label{eq:total-training-objective}
\end{equation}
where $\mathcal{L}_{\mathrm{NTP}}$ trains the model for next-token prediction, $\mathcal{L}_{\mathrm{NCP}}$ trains the Concept Module and Token Encoder for next-concept prediction, and $\mathcal{L}_{\mathrm{VQ}}$ fits the codebook entries to the continuous concept representations. The coefficients $\alpha$ and $\beta$ control the weights of the two auxiliary objectives. The three objectives are jointly optimized during training.

\subsection{Optimization}
\label{subsec:optimization}

We use Moonlight Muon~\citep{liu2025muonscalablellmtraining}, built on the Muon optimizer~\citep{jordan2024muon}, for the matrix-valued parameters.  For a matrix parameter $\mathbf{W}\in\mathbb{R}^{d_{\mathrm{out}}\times d_{\mathrm{in}}}$, the update can be written as
\begin{equation}
    \mathbf{W}_{t}
    =\mathbf{W}_{t-1}
     -\eta_t\left(
        \lambda_{Muon}\,\mathbf{O}_{t}
        \sqrt{\max(d_{\mathrm{in}},d_{\mathrm{out}})}
        +\lambda_{wd}\mathbf{W}_{t-1}
      \right),
    \label{eq:muon-update}
\end{equation}
where $\mathbf{O}_t$ is the orthogonalized Muon update, $\eta_t$ is the learning-rate schedule, $\lambda_{Muon}$ controls the update scale, and $\lambda_{wd}$ denotes weight decay. Our default setting uses a learning rate of $6\times10^{-5}$ and the same cosine scheduler as Olmo-3-7B. Embeddings, biases, and other non-Muon parameters are optimized with AdamW; other parameters are optimized by Muon.

\section{Experiments}
\label{sec:evaluation}

\subsection{Experimental Setup}

\textbf{Models and Datasets: }
We use OLMo-3-7B as the backbone of \texttt{NCP-ArchPreview} and follow the staged data curriculum of OLMo-3. The first training stage uses Dolma 3 Mix, and the second stage continues training on Dolma 3 Dolmino. We evaluate the resulting models following the OLMo evaluation protocol implemented in OLMo-Core, which covers 30 benchmark families, including MMLU~\citep{hendrycks2021mmlu}, GSM8K~\citep{cobbe2021gsm8k}, MATH-500~\citep{hendrycks2021math,lightman2023verify}, HumanEval~\citep{chen2021humaneval}, MBPP~\citep{austin2021mbpp}, ARC~\citep{clark2018arc}, and HellaSwag~\citep{zellers2019hellaswag}. These benchmarks evaluate factual knowledge, mathematical reasoning, code generation, commonsense reasoning, reading comprehension, and language modeling. The complete task list, prompting configurations, and evaluation metrics are provided in Appendix~\ref{app:evaluation-details}.

For the VQ training study in Section~\ref{sec:vq-training}, we continue training the Stage-1 checkpoint separately on Magicoder~\citep{wei2023magicoder}, Orca-Math~\citep{mitra2024orca}, and TriviaQA-RC~\citep{joshi2017triviaqa} for code, mathematics, and knowledge adaptation, respectively. Target-domain performance is evaluated using HumanEval and MBPP with their EvalPlus variants for code~\citep{chen2021humaneval,austin2021mbpp}, GSM8K and MATH-500 for mathematics~\citep{cobbe2021gsm8k,hendrycks2021math,lightman2023verify}, and TriviaQA for knowledge~\citep{joshi2017triviaqa}.

\textbf{Configuration for \texttt{NCP-ArchPreview}: }
\texttt{NCP-ArchPreview} retains the main configuration of OLMo-3-7B, including a hidden size of 4,096, an FFN size of 11,008, and 32 attention heads. Its 32 token-level Transformer layers are divided evenly into a 16-layer Token Encoder and a 16-layer Token Decoder. The Concept Module consists of 8 Transformer layers and uses a chunk size of 4 tokens. The VQ module contains 32 codebooks, each with 128 entries of dimension 128. Concept representations are incorporated through normalized residual connections at every decoder layer. The training objective combines the standard next-token prediction loss with the VQ and NCP losses. The resulting model contains approximately 8.94 billion parameters and uses a maximum context length of 8,192 tokens.

\subsection{Main Results}

\subsubsection{Overview Across Training Stages}
We compare \texttt{NCP-ArchPreview} with OLMo-3-7B across the first two stages of training. Figure~\ref{fig:stage-loss} compares their token-level language-modeling losses, while Table~\ref{tab:core88-results} summarizes downstream performance under the same evaluation protocol. Overall, \texttt{NCP-ArchPreview} achieves a lower language-modeling loss and a higher macro-average over the downstream metrics in both stages.

\begin{figure}[tbp]
\centering
\includegraphics[width=\linewidth]{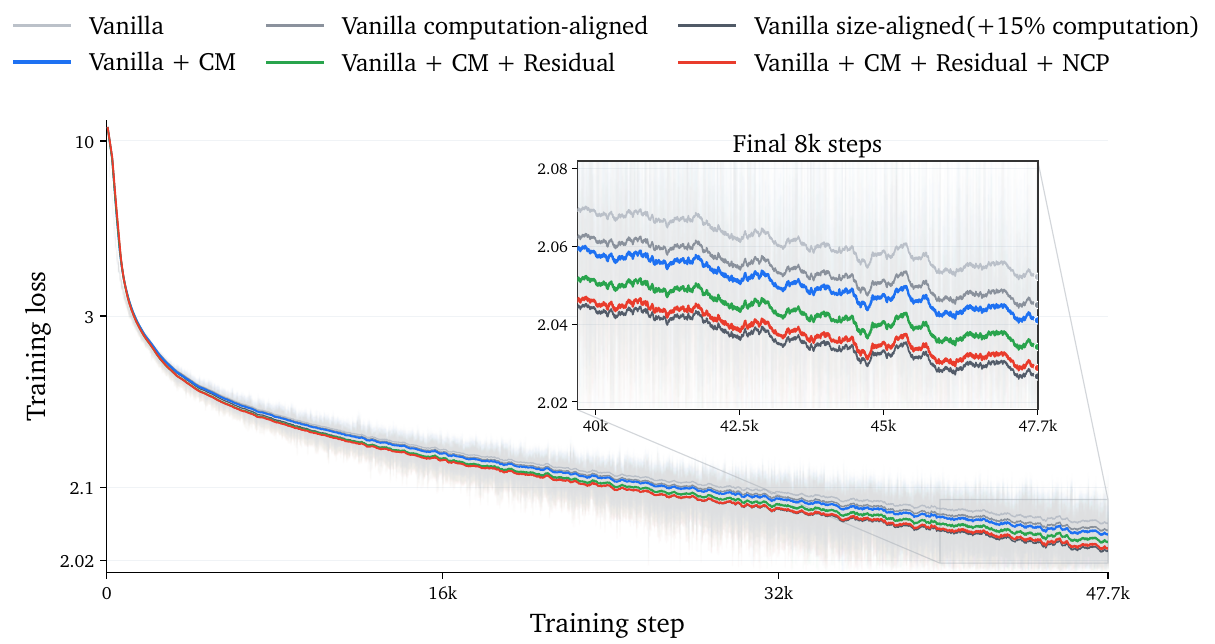}
\caption{Training-loss ablation of \texttt{NCP-ArchPreview} over the first 200B training tokens. All runs use the same training recipe. The six curves correspond to three standard OLMo-3-7B baselines \texttt{Vanilla}, \texttt{Vanilla computation-aligned}, and \texttt{Vanilla size-aligned} (+15\% computation) and three progressive configurations \texttt{Vanilla+CM}, \texttt{Vanilla+CM+Residual}, and \texttt{Vanilla+CM+Residual+NCP}, with the last corresponding to \texttt{NCP-ArchPreview}. The inset enlarges the final 8k training steps to better resolve the differences among the configurations.}
\label{fig:ablation}
\end{figure}

\def\maingain#1{\,{\raisebox{-0.35ex}{\scriptsize\textcolor{green!60!black}{#1}}}}
\def\maindrop#1{\,{\raisebox{-0.35ex}{\scriptsize\textcolor{red!80!}{#1}}}}
\begin{table}[tbp]
    \centering
    \small
    \setlength{\tabcolsep}{2.2pt}
    \renewcommand{\arraystretch}{0.98}
    \caption{Downstream performance grouped by task domain. Vanilla denotes the released OLMo-3 model for each stage. Metrics above the likelihood block are reported as percentages and are higher-is-better. The likelihood block reports bits per UTF-8 byte (BPB), where lower is better. The BPB average is computed separately and is excluded from Overall AVG.}
    \label{tab:core88-results}

    \begin{tabular}{llcccc}
        \toprule
        \multirow{2}{*}{Category}
        & \multirow{2}{*}{Dataset}
        & \multicolumn{2}{c}{Stage-1}
        & \multicolumn{2}{c}{Stage-2} \\
        & & Vanilla & NCP
        & Vanilla & NCP \\
        \midrule

       \multirow{7}{*}{MMLU} &
       MMLU-Humanities
       & 64.73 & 68.28\maingain{+3.55}
       & 69.29 & 70.78\maingain{+1.49} \\
       & MMLU-Social Sciences
       & 71.18 & 73.97\maingain{+2.79}
       & 74.27 & 76.79\maingain{+2.52} \\
       & MMLU-STEM
       & 53.04 & 55.53\maingain{+2.49}
       & 59.89 & 61.01\maingain{+1.12} \\
       & MMLU-Other
       & 64.00 & 65.63\maingain{+1.63}
       & 66.40 & 68.84\maingain{+2.44} \\
       & MMLU-Pro
       & 19.53 & 20.26\maingain{+0.73}
       & 21.77 & 22.29\maingain{+0.52} \\
       & \textbf{MMLU}
       & \textbf{62.22} & \textbf{64.80}\maingain{\textbf{+2.58}}
       & \textbf{66.66} & \textbf{68.48}\maingain{\textbf{+1.82}} \\
       & \textbf{AVG}
       & \textbf{54.50} & \textbf{56.73}\maingain{\textbf{+2.24}}
       & \textbf{58.32} & \textbf{59.94}\maingain{\textbf{+1.62}} \\
       \midrule[\heavyrulewidth]

        \multirow{5}{*}{MATH}
        & GSM8K
        & 39.27 & 45.26\maingain{+5.99}
        & 79.68 & 83.02\maingain{+3.34} \\
        & GSM-Symbolic
        & 18.85 & 22.80\maingain{+3.95}
        & 57.32 & 60.32\maingain{+3.00} \\
        & Minerva
        & 12.52 & 15.63\maingain{+3.11}
        & 42.09 & 42.32\maingain{+0.23} \\
        & MATH-500
        & 12.52 & 14.48\maingain{+1.96}
        & 43.44 & 43.91\maingain{+0.47} \\
        & \textbf{AVG}
        & \textbf{20.79} & \textbf{24.54}\maingain{\textbf{+3.75}}
        & \textbf{55.63} & \textbf{57.39}\maingain{\textbf{+1.76}} \\
        \midrule[\heavyrulewidth]

        \multirow{7}{*}{Code}
        & BigCodeBench
        & 20.68 & 22.79\maingain{+2.11}
        & 39.58 & 38.42\maindrop{-1.16} \\
        & HumanEval
        & 27.10 & 31.38\maingain{+4.28}
        & 49.31 & 45.62\maindrop{-3.69} \\
        & DS-1000
        & 18.58 & 20.40\maingain{+1.82}
        & 25.10 & 26.00\maingain{+0.90} \\
        & MBPP
        & 34.53 & 35.91\maingain{+1.38}
        & 48.98 & 50.85\maingain{+1.87} \\
        & MultiPL-E HumanEval
        & 21.34 & 22.27\maingain{+0.93}
        & 31.92 & 30.90\maindrop{-1.02} \\
        & MultiPL-E MBPP
        & 28.67 & 33.96\maingain{+5.29}
        & 41.64 & 40.85\maindrop{-0.79} \\
        & \textbf{AVG}
        & \textbf{25.15} & \textbf{27.79}\maingain{\textbf{+2.64}}
        & \textbf{39.42} & \textbf{38.77}\maindrop{\textbf{-0.65}} \\
        \midrule[\heavyrulewidth]

        \multirow{3}{*}{MC-STEM}
        & ARC-E
        & 90.95 & 92.30\maingain{+1.35}
        & 93.81 & 94.07\maingain{+0.26} \\
        & ARC-C
        & 77.99 & 81.57\maingain{+3.58}
        & 85.49 & 83.28\maindrop{-2.21} \\
        & \textbf{AVG}
        & \textbf{84.47} & \textbf{86.93}\maingain{\textbf{+2.47}}
        & \textbf{89.65} & \textbf{88.67}\maindrop{\textbf{-0.98}} \\
        \midrule[\heavyrulewidth]

        \multirow{4}{*}{MC-Non-STEM}
       & PiQA
       & 72.25 & 80.85\maingain{+8.60}
       & 78.35 & 81.45\maingain{+3.10} \\
       & CommonsenseQA
       & 72.65 & 74.04\maingain{+1.39}
       & 78.05 & 77.48\maindrop{-0.57} \\
       & SocialIQA
       & 65.35 & 69.24\maingain{+3.89}
       & 74.16 & 74.36\maingain{+0.20} \\
       & \textbf{AVG}
       & \textbf{70.08} & \textbf{74.71}\maingain{\textbf{+4.63}}
       & \textbf{76.85} & \textbf{77.76}\maingain{\textbf{+0.91}} \\
       \midrule[\heavyrulewidth]

       \multirow{7}{*}{GenQA}
       & HellaSwag
       & 66.05 & 66.05\maingain{+0.00}
       & 65.60 & 66.40\maingain{+0.80} \\
       & WinoGrande
       & 49.17 & 51.22\maingain{+2.05}
       & 50.04 & 49.64\maindrop{-0.40} \\
       & LAMBADA-Standard
       & 66.60 & 66.89\maingain{+0.29}
       & 65.55 & 67.13\maingain{+1.58} \\
       & LAMBADA-OpenAI
       & 72.13 & 72.21\maingain{+0.08}
       & 69.22 & 69.96\maingain{+0.74} \\
       & MedMCQA
       & 32.98 & 33.21\maingain{+0.23}
       & 32.36 & 32.99\maingain{+0.63} \\
       & MedQA
       & 38.81 & 38.96\maingain{+0.15}
       & 38.18 & 38.10\maindrop{-0.08} \\
       & \textbf{AVG}
       & \textbf{54.29} & \textbf{54.76}\maingain{\textbf{+0.47}}
       & \textbf{53.49} & \textbf{54.04}\maingain{\textbf{+0.55}} \\

       \midrule[\heavyrulewidth]
       \textbf{Overall AVG} &
       & \textbf{46.59} & \textbf{49.04}\maingain{\textbf{+2.45}}
       & \textbf{56.98} & \textbf{57.57}\maingain{\textbf{+0.59}} \\

        \midrule[\heavyrulewidth]
        \multirow{11}{*}{Likelihood($\downarrow$)}
        & HumanEval Gold
        & 0.384 & 0.364\maingain{-0.020}
        & 0.362 & 0.339\maingain{-0.023} \\
        & MBPP Gold
        & 0.475 & 0.473\maingain{-0.002}
        & 0.443 & 0.454\maindrop{+0.011} \\
        & MT-MBPP Gold
        & 0.437 & 0.434\maingain{-0.004}
        & 0.402 & 0.426\maindrop{+0.025} \\
        & CoQA
        & 0.366 & 0.370\maindrop{+0.004}
        & 0.305 & 0.345\maindrop{+0.040} \\
        & DROP
        & 4.474 & 4.397\maingain{-0.077}
        & 4.431 & 4.146\maingain{-0.285} \\
        & GSM8K Gold
        & 0.405 & 0.400\maingain{-0.005}
        & 0.356 & 0.381\maindrop{+0.024} \\
        & Jeopardy
        & 0.352 & 0.347\maingain{-0.005}
        & 0.385 & 0.358\maingain{-0.027} \\
        & LAMBADA-OpenAI
        & 0.295 & 0.288\maingain{-0.007}
        & 0.320 & 0.307\maingain{-0.014} \\
        & Natural Questions
        & 0.917 & 0.884\maingain{-0.033}
        & 0.841 & 0.777\maingain{-0.064} \\
        & SQuAD
        & 0.140 & 0.151\maindrop{+0.011}
        & 0.083 & 0.098\maindrop{+0.015} \\
        & \textbf{AVG}
        & \textbf{0.824} & \textbf{0.811}\maingain{\textbf{-0.014}}
        & \textbf{0.793} & \textbf{0.763}\maingain{\textbf{-0.030}} \\
       
       \bottomrule
    \end{tabular}
\end{table}

\subsubsection{Training Loss Performance}

\texttt{NCP-ArchPreview} exhibits lower training loss than OLMo-3-7B throughout Stage-1 and Stage-2, as shown in Figure~\ref{fig:stage-loss}. Both models are trained on the same data. During Stage-1, the loss gap between \texttt{NCP-ArchPreview} and OLMo-3-7B widens as training progresses, reaching 0.091 by the end of the stage. \texttt{NCP-ArchPreview} reaches the final loss of OLMo-3-7B after consuming only 51.3\% of the training tokens, corresponding to 1.95$\times$ faster convergence in terms of training tokens. The optimization advantage remains evident in Stage-2: \texttt{NCP-ArchPreview} achieves a 0.027 lower final loss while matching the final loss of OLMo-3-7B with only 66.2\% of the training tokens, yielding a 1.51$\times$ convergence speedup.

\subsubsection{Downstream Performance}

\texttt{NCP-ArchPreview} performs better than OLMo-3-7B on downstream tasks at both stages. At Stage-1, \texttt{NCP-ArchPreview} outperforms OLMo-3-7B on almost all evaluated benchmarks, with an average score that is 2.45 points higher across all evaluated datasets. The improvements are particularly large on MATH, Code, and MC-Non-STEM. Among them, the largest improvement is on MATH, reaching 18\%. This is consistent with the lower training loss we observe. 

At Stage-2, \texttt{NCP-ArchPreview} achieves a 0.59-point higher macro-average, with improvements on MATH, MC-Non-STEM, and GenQA. However, Stage-2 contains only approximately 10\% code data. Better fitting of the aggregate mixture can therefore improve the overall average while weakening performance on underrepresented domains such as code.

\subsubsection{Relationship Between Training Loss and Downstream Performance}

We examine whether training loss at different training stages accurately reflects model capability. At Stage-1, we evaluate checkpoints saved at different points during training on the same downstream tasks (See Appendix~\ref{app:evaluation-details} for detailed downstream scores). As training progresses, the loss gradually decreases, and the downstream scores continuously increase, indicating that model capability improves throughout training. At Stage-2, we experiment with three training configurations, V1--V3. Although their final training losses become progressively lower, their downstream performance becomes progressively worse. We believe that this phenomenon may result from a mismatch between the Stage-2 training data distribution and the downstream-task data distribution. We discuss this issue further with a proxy dataset in Appendix~\ref{sec:mid-training-data-proxy}.

\subsection{Ablation Studies}

We ablate the key components of \texttt{NCP-ArchPreview} and conduct parameter- and computation-aligned comparisons against standard OLMo-3. Specifically, we first compare \texttt{NCP-ArchPreview} with size-aligned and computation-aligned standard OLMo-3-7B models, and then examine the contributions of the Concept Module, residual connections, and the NCP loss through progressive component ablations.

\subsubsection{Matched Model Size and Computation}
\label{sec:matched_model_size_and_computation}
To evaluate whether the improvement of \texttt{NCP-ArchPreview} comes from the proposed architecture rather than simply from additional parameters or computation, we compare it with three OLMo-3-based baselines under the same experimental settings: \texttt{Vanilla}, \texttt{Vanilla size-aligned}, and \texttt{Vanilla computation-aligned}.

As described in the architecture section, the Concept Module groups every four token-level hidden states into one concept representation, reducing the Concept Module sequence length to approximately one quarter of the original sequence. Thus, each Concept Module block introduces roughly the same number of parameters as a standard OLMo-3-7B block while requiring only about one quarter of its computation.

For clarity, we denote the parameter count and computation of one standard OLMo-3-7B Transformer block as $P_{\mathrm{blk}}$ and $F_{\mathrm{blk}}$, respectively. Accordingly, one Concept Module block contributes approximately $P_{\mathrm{blk}}$ parameters but less than one quarter of its analytical training FLOPs. Since the VQ codebook and residual components account for only a negligible fraction of the total parameters and computation, we omit them when constructing the matched baselines. Detailed model configurations are shown in Table~\ref{tab:ab-model_size_computation}.

\begin{table}[tbp]
\centering
\small
\caption{Model configurations under parameter- and computation-aligned settings. $P_{\mathrm{blk}}$ and $F_{\mathrm{blk}}$ denote the parameter count and computation of one standard OLMo-3-7B Transformer block, respectively. The \texttt{Vanilla size-aligned} baseline uses 40 standard OLMo-3-7B blocks to match the approximate parameter count of \texttt{NCP-ArchPreview}, while the \texttt{Vanilla computation-aligned} baseline uses 34 blocks to match its approximate computation.}
\label{tab:ab-model_size_computation}
\begin{tabular}{@{}cccccc@{}}
\toprule
Model & \makecell{Token \\ Encoder} & \makecell{Token \\ Decoder} & \makecell{Concept \\ Module} & Parameters & Computation \\
\midrule
\texttt{NCP-ArchPreview}
& $16P_{\mathrm{blk}}$
& $16P_{\mathrm{blk}}$
& $8P_{\mathrm{blk}}$
& $40P_{\mathrm{blk}}$
& $34F_{\mathrm{blk}}$ \\
\texttt{Vanilla}
& $32P_{\mathrm{blk}}$
& -- & --
& $32P_{\mathrm{blk}}$
& $32F_{\mathrm{blk}}$ \\
\texttt{Vanilla size-aligned}
& $40P_{\mathrm{blk}}$
& -- & --
& $40P_{\mathrm{blk}}$
& $40F_{\mathrm{blk}}$ \\
\texttt{Vanilla computation-aligned}
& $34P_{\mathrm{blk}}$
& -- & --
& $34P_{\mathrm{blk}}$
& $34F_{\mathrm{blk}}$ \\
\bottomrule
\end{tabular}
\end{table}

As shown in Figure~\ref{fig:ablation}, \texttt{NCP-ArchPreview} substantially outperforms both \texttt{Vanilla} and \texttt{Vanilla computation-aligned}, indicating that its improvement cannot be explained solely by additional computation. Moreover, \texttt{NCP-ArchPreview} approaches the performance of \texttt{Vanilla size-aligned}, despite using only $34/40=85\%$ of its computation.

\subsubsection{Module Ablation}

We further analyze the contributions of the key components in \texttt{NCP-ArchPreview} by progressively introducing them into the standard OLMo-3-7B architecture. Specifically, we compare \texttt{Vanilla}, \texttt{Vanilla+CM}, \texttt{Vanilla+CM+Residual}, and \texttt{Vanilla+CM +Residual+NCP}, where \texttt{CM} denotes the Concept Module
and the final configuration corresponds to \texttt{NCP-ArchPreview}.

As shown in Figure~\ref{fig:ablation}, introducing the Concept Module improves the training loss over \texttt{Vanilla}. Adding the residual component further reduces the loss, and incorporating the NCP loss yields an additional improvement. These results show that the components of \texttt{NCP-ArchPreview} progressively improve model performance, with the complete model approaching the \texttt{Vanilla size-aligned} baseline.

\subsection{Hierarchical Residual Connection Comparison}
\label{subsec:residual-connection-comparison}

We evaluate variants of the hierarchical residual connections on the 1B-scale \texttt{NCP-ArchPreview} model after 150B training tokens. The reported loss is averaged over the final 200 logged steps. The model without hierarchical residual connections serves as the common reference for both loss and computation. We report the loss difference and additional analytical training FLOPs relative to this reference.

Input-Level Cross-Module Connections augment Intra-Module Residual Connections (IRC) with three embedding-level connections: Token Encoder embedding $\rightarrow$ Concept Module embedding, Concept Module embedding $\rightarrow$ Token Decoder embedding, and Token Encoder embedding $\rightarrow$ Token Decoder embedding. Block AttnRes~\citep{kimiteam2026attentionresiduals} groups residual updates into blocks of four and also includes cross-module connections. Table~\ref{tab:residual-connection-ablation} compares these variants with IRC and the complete hierarchical residual connection design.

IRC + CRC achieves the largest loss reduction, improving over the reference without hierarchical residual connections by 0.0323 with an additional 0.051\% analytical training FLOPs. IRC alone achieves a 0.0273 loss reduction with only 0.024\% additional FLOPs. Adding the three input-level cross-module connections further improves the loss by 0.0023, achieving a 0.0296 reduction while introducing no additional counted tensor contractions. Applying softmax normalization at all connection stages gives
a similar result, with a loss that is only 0.0003 higher at the same analytical FLOPs.

The Block AttnRes variant includes cross-module connections from the Token Encoder to the Concept Module and from the Token Encoder and Concept Module to the Token Decoder. With a comparable 0.026\% FLOPs increase, it achieves a 0.0180 loss reduction, which is smaller than the improvement of the IRC-based variants. These analytical FLOPs do not account for source-state materialization, memory traffic, reduction operations, or small-kernel launch overhead. The complete hierarchical connection design may therefore incur additional runtime and memory costs beyond those reflected in the analytical FLOPs.

\begin{table}[tbp]
    \centering
    \caption{Ablation of hierarchical residual connections on the 1B-scale \texttt{NCP-ArchPreview} model after 150B training tokens. Loss is averaged over the final 200 logged steps. Negative loss differences indicate improvements over the reference without hierarchical residual connections; positive FLOPs differences indicate additional analytical training computation.}
    \label{tab:residual-connection-ablation}
    \small
    \setlength{\tabcolsep}{4pt}
    \renewcommand{\arraystretch}{1.12}
    \begin{tabular}{@{}p{0.6\linewidth}ccc@{}}
        \toprule
        Variant & \shortstack{Avg. LM\\loss}
        & \shortstack{$\Delta$ loss\\vs. reference}
        & \shortstack{$\Delta$ FLOPs\\vs. reference} \\
        \midrule
        NCP-ArchPreview
        (without hierarchical residual connections)
        & 2.2588 & 0.0000 & $0.000\%$ \\
        IRC + CRC
        & 2.2265 & $-0.0323$ & $+0.051\%$ \\
        \midrule
        IRC only
        & 2.2315 & $-0.0273$ & $+0.024\%$ \\
        IRC + Input-Level Cross-Module Connections
        & 2.2292 & $-0.0296$ & $+0.024\%$ \\
        IRC + All-Stage Softmax
        & 2.2295 & $-0.0294$ & $+0.024\%$ \\
        \midrule
        Block AttnRes ($S=4$) + Cross-Module Connections
        & 2.2409 & $-0.0180$ & $+0.026\%$ \\
        \bottomrule
    \end{tabular}
\end{table}

\subsection{Scaling Law}
\label{sec:scaling-law}
\begin{figure}[tbp]
    \centering
    \includegraphics[width=0.82\linewidth]{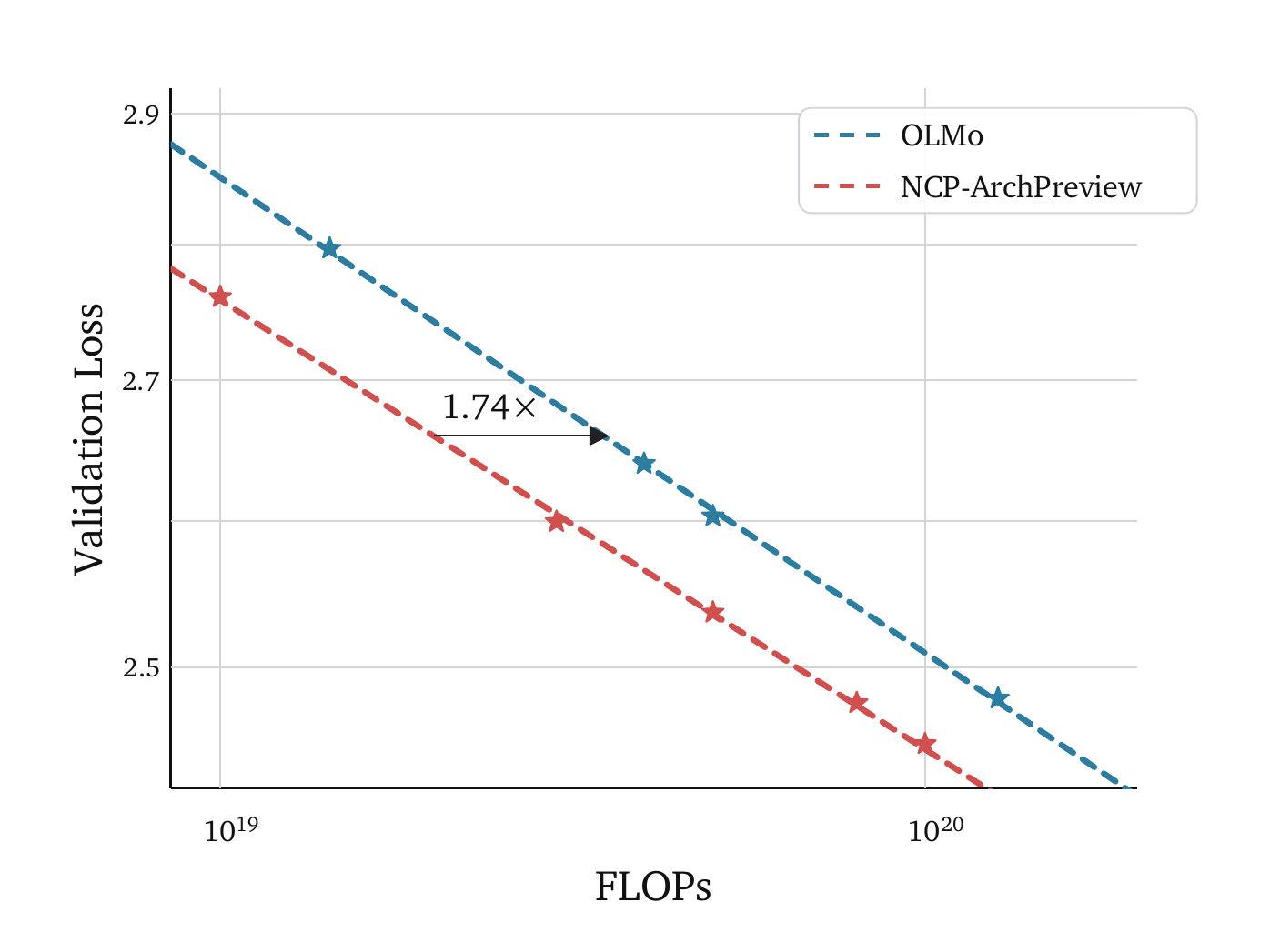}
    \caption{The scaling law curves for OLMo-3 and \texttt{NCP-ArchPreview}. Each marker denotes the best validation loss found
    after searching training hyperparameters and model/data allocations at a
    fixed FLOPs budget. Dashed curves show the corresponding scaling fits.}
    \label{fig:scaling-law-nd}
\end{figure}

For comparison with OLMo-3  as the compute budget increases, we conduct a series of scaling experiments at multiple FLOPs budgets~\citep{olmo2026olmo3}. For each budget, we search over training hyperparameters and the allocation between model size and training data, then report the best
validation loss found at that budget. As shown in Figure~\ref{fig:scaling-law-nd}, \texttt{NCP-ArchPreview} achieves  1.74x computational efficiency compared to OLMo-3 with compute-optimal training. The details are described in Appendix~\ref{app:scaling-experiments}.

\subsection{Model Analysis}

\subsubsection{Numerical Stability of the OLMo-3-7B Configuration}

When training \texttt{NCP-ArchPreview} with the layer-wise Q/K normalization adopted by OLMo-3, we observe persistent growth of attention logits. A pronounced imbalance in the matrix norms of the Q/K head blocks accompanies this growth. A small number of outlier heads can dominate the pre-softmax scores, leading to highly concentrated attention distributions. The enlarged Q/K states are also associated with uneven gradient flow, abnormal Q/K gradient norms, and eventual spikes in the global gradient norm.

\begin{figure}[tbp]
    \centering
    \includegraphics[width=0.85\linewidth]{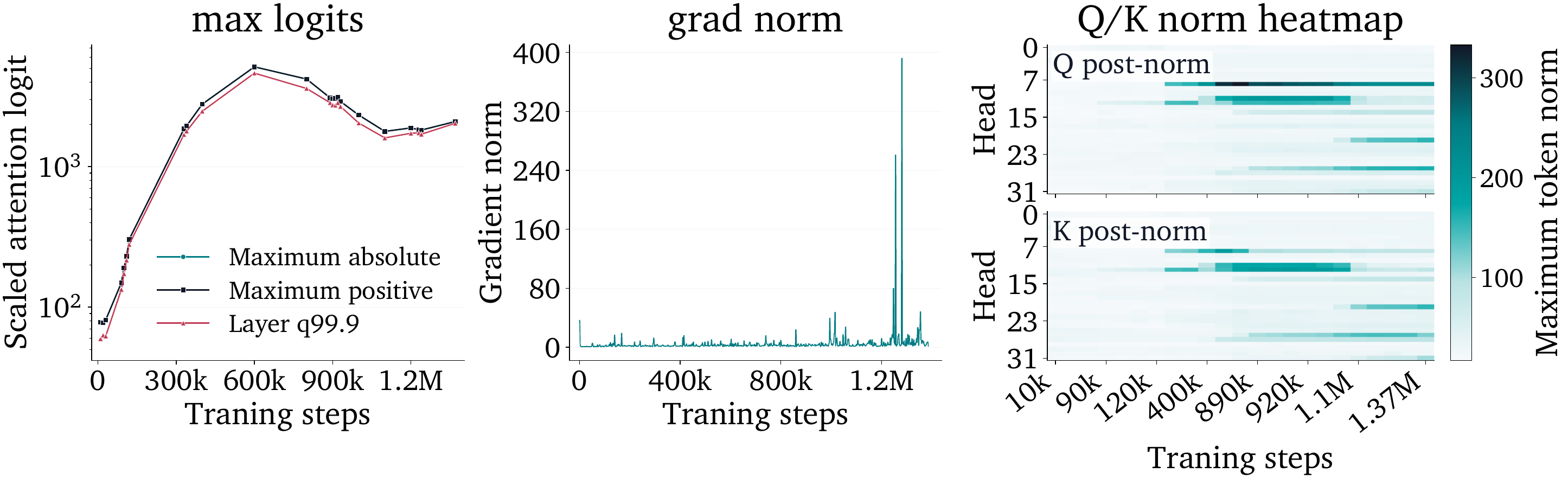}
    
    \caption{Numerical instabilities observed when using the OLMo-3-7B layer-wise Q/K normalization. Exploding attention logits (left) coincide with severe gradient-norm spikes (middle) and head-wise outliers in the post-normalization Q/K norms (right).}
\end{figure}

\subsubsection{Ablation of the Instability}

To identify the source of this behavior, we compare matched AdamW and Muon baselines and evaluate Muon variants replacing layer-wise Q/K normalization with per-head Q/K normalization. All other training settings remain unchanged. The Muon baseline already reproduces the instability. In contrast, \texttt{NCP-ArchPreview} performs better in terms of stability. Per-head Q/K normalization consistently suppresses both Q/K norm outliers and attention-logit growth.

These results suggest that the primary source of the instability is the interaction between full-matrix Muon updates and layer-wise Q/K normalization. Full-matrix Muon couples the updates of multiple attention heads, whereas layer-wise normalization constrains only their aggregate scale. It does not independently control the scale or QK alignment of each head, allowing a small number of heads to dominate and produce large dot products progressively. This interpretation is consistent with prior observations that Muon can be susceptible to attention-logit explosion concentrated in a subset of heads~\citep{kimiteam2026kimik3openfrontier}.

\begin{figure}[tbp]
    \centering
    \includegraphics[width=\linewidth]{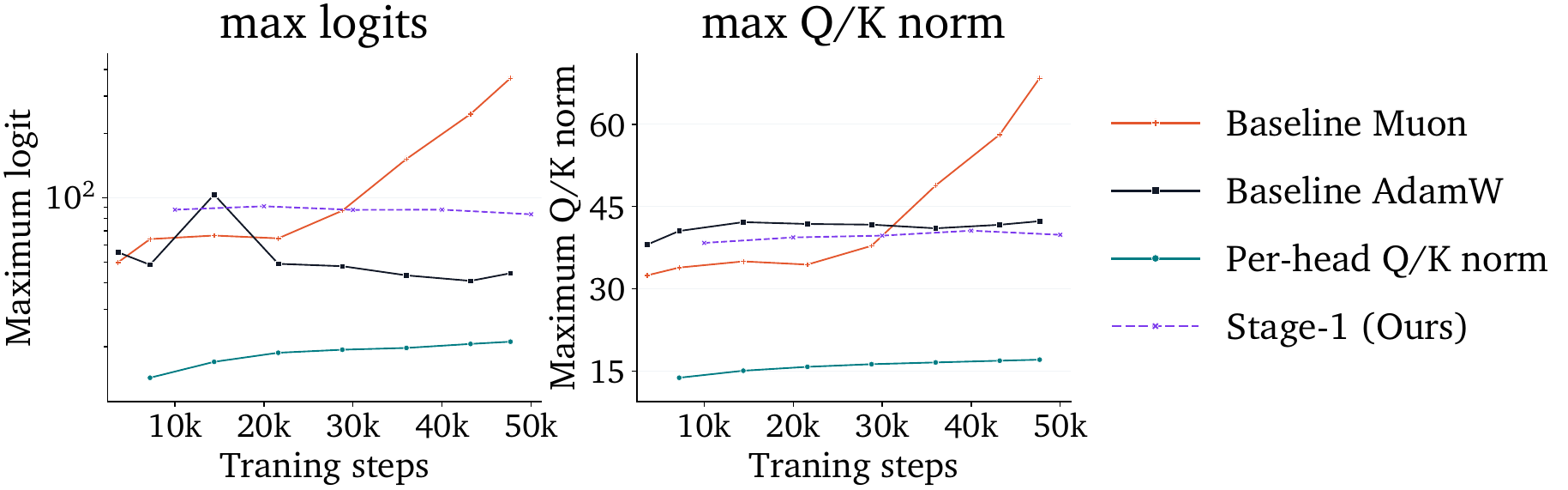}
    \caption{Ablation of the numerical instability. The maximum absolute attention logit (left) and post-normalization Q/K norm (right) grow under the Muon baseline, whereas per-head Q/K normalization keeps both bounded.}
\end{figure}

Per-head Q/K normalization provides a more effective stabilization method in this analysis. Nevertheless, we use the layer-wise Q/K normalization adopted by OLMo-3-7B in all main experiments to preserve a controlled comparison with the original baseline. The per-head variant is reported only as a stabilization intervention in the ablation study, and is not used to obtain the main performance results.

\section{Additional Results}

\subsection{VQ Training}
\label{sec:vq-training}

The discrete concept vocabulary provides \texttt{NCP-ArchPreview} with a compact interface between token representations and concept-level prediction. We investigate whether this interface can also support lightweight domain adaptation. All adaptation experiments start from the corresponding Stage-1 checkpoints of OLMo-3-7B and \texttt{NCP-ArchPreview}.

For \texttt{NCP-ArchPreview}, the \texttt{+VQ} setting freezes the token-level backbone and updates only the VQ codebooks and concept-prediction heads, corresponding to 17M trainable parameters. We compare this setting with full-parameter training (\texttt{+Full}) and low-rank adaptation (\texttt{+LoRA})~\citep{hu2021loralowrankadaptationlarge}. To provide a parameter-matched comparison, we configure LoRA with 17M trainable parameters. We also apply Full and LoRA training to OLMo-3-7B using the same domain-specific data and training settings. Table~\ref{tab:vq-parameter-cost} summarizes the parameter costs. The \texttt{+VQ} and \texttt{+LoRA} settings each update 17M parameters, while \texttt{+Full} updates the entire model. VQ adaptation introduces no new parameters because it updates existing codebooks and prediction heads, whereas LoRA adds 17M parameters to the model.

\begin{table}[tbp]
\centering
\caption{Parameters of the three training strategies. VQ and Full training
update existing weights without adding parameters, whereas LoRA introduces
more parameters.}
\label{tab:vq-parameter-cost}
\begingroup
\small
\setlength{\tabcolsep}{14pt}
\renewcommand{\arraystretch}{1.08}
\begin{tabular}{@{}lcc@{}}
\toprule
Method &
\makecell{Trainable \\ Parameters} &
\makecell{Added \\ Parameters} \\
\midrule
+Full & 8.9B & 0 \\
+LoRA & 17M & 17M \\
+VQ & 17M & 0 \\
\bottomrule
\end{tabular}
\endgroup
\end{table}

\paragraph{Code:}

Table~\ref{tab:vq-code-results} reports adaptation on code benchmarks. For \texttt{NCP-ArchPreview}, VQ training raises the code average from 30.04 to 32.69 (+2.65), the highest among the adapted variants, and is the only setting that improves all four code tasks. Full training improves HumanEval, HumanEval+, and MBPP, but its 22.49-point drop on MBPP+ lowers the aggregate below its Stage-1 baseline. LoRA also loses 7.67 points on MBPP+. Against the matched OLMo-3-7B controls, the Full and LoRA variants of \texttt{NCP-ArchPreview} achieve higher code averages and higher general averages. Such post-adaptation regressions are consistent with interference documented in domain adaptation and continual learning, where optimization on new data can reduce previously acquired
capabilities~\citep{ICLR2024_d51cd79a,Kirkpatrick_2017,van_de_Ven_2025}. Although VQ training leaves the token-level backbone unchanged, limiting interference with previously learned code capabilities, its general average nevertheless decreases by 0.42 points.

\begin{table}[tbp]
\centering
\caption{Code adaptation from Stage-1 across four programming benchmarks,
together with performance retained on general benchmarks.}
\label{tab:vq-code-results}
\begingroup
\def\vqgain#1{\,{\raisebox{-0.35ex}{\scriptsize\textcolor{green!60!black}{#1}}}}
\def\vqdrop#1{\,{\raisebox{-0.35ex}{\scriptsize\textcolor{red!80!}{#1}}}}
\def\vqscore#1#2{#1#2}
\footnotesize
\setlength{\tabcolsep}{2.6pt}
\renewcommand{\arraystretch}{1.04}

\begin{adjustbox}{max width=\linewidth,center}
\begin{tabular}{@{}lccccc@{\hspace{6pt}}cccccc@{}}
\toprule
\multirow{2}{*}{\raisebox{-0.85ex}{\textbf{Model}}} &
\multicolumn{5}{c}{\textbf{Code}} &
\multicolumn{6}{c}{\textbf{General}} \\
\cmidrule(lr){2-6}\cmidrule(lr){7-12}
& HumanEval & HumanEval+ & MBPP & MBPP+ & \textbf{Avg.}
& MMLU & ARC-E & HellaSwag & NQ Open & SciQ MC & \textbf{Avg.} \\
\midrule
OLMo-3-7B
& 26.94 & 23.17 & 34.27 & 28.04 & 28.11
& 62.25 & 91.08 & 66.25 & 29.46 & 88.50 & 67.51 \\
+Full
& \vqscore{\textbf{53.35}}{\vqgain{\textbf{+26.41}}}
& \vqscore{15.85}{\vqdrop{-7.32}}
& \vqscore{\textbf{38.88}}{\vqgain{\textbf{+4.61}}}
& \vqscore{2.38}{\vqdrop{-25.66}}
& \vqscore{\textbf{27.62}}{\vqdrop{\textbf{-0.49}}}
& \vqscore{\textbf{61.43}}{\vqdrop{\textbf{-0.82}}}
& \vqscore{90.70}{\vqdrop{-0.38}}
& \vqscore{\textbf{65.85}}{\vqdrop{\textbf{-0.40}}}
& \vqscore{\textbf{28.22}}{\vqdrop{\textbf{-1.24}}}
& \vqscore{88.20}{\vqdrop{-0.30}}
& \vqscore{\textbf{66.88}}{\vqdrop{\textbf{-0.63}}} \\
+LoRA
& \vqscore{34.05}{\vqgain{+7.11}}
& \vqscore{\textbf{19.51}}{\vqdrop{\textbf{-3.66}}}
& \vqscore{37.81}{\vqgain{+3.54}}
& \vqscore{\textbf{6.35}}{\vqdrop{\textbf{-21.69}}}
& \vqscore{24.43}{\vqdrop{-3.68}}
& \vqscore{61.13}{\vqdrop{-1.12}}
& \vqscore{\textbf{91.12}}{\vqgain{\textbf{+0.04}}}
& \vqscore{65.15}{\vqdrop{-1.10}}
& \vqscore{28.06}{\vqdrop{-1.40}}
& \vqscore{\textbf{88.70}}{\vqgain{\textbf{+0.20}}}
& \vqscore{66.83}{\vqdrop{-0.68}} \\
\midrule
NCP-ArchPreview
& 31.27 & 23.17 & 36.09 & 29.63 & 30.04
& 64.77 & 92.89 & 65.85 & 32.10 & 87.80 & 68.68 \\
+Full
& \vqscore{\textbf{43.27}}{\vqgain{\textbf{+12.00}}}
& \vqscore{23.78}{\vqgain{+0.61}}
& \vqscore{\textbf{39.53}}{\vqgain{\textbf{+3.44}}}
& \vqscore{7.14}{\vqdrop{-22.49}}
& \vqscore{28.43}{\vqdrop{-1.61}}
& \vqscore{63.24}{\vqdrop{-1.53}}
& \vqscore{\textbf{92.76}}{\vqdrop{\textbf{-0.13}}}
& \vqscore{65.55}{\vqdrop{-0.30}}
& \vqscore{30.15}{\vqdrop{-1.95}}
& \vqscore{\textbf{89.30}}{\vqgain{\textbf{+1.50}}}
& \vqscore{68.20}{\vqdrop{-0.48}} \\
+LoRA
& \vqscore{36.49}{\vqgain{+5.22}}
& \vqscore{28.05}{\vqgain{+4.88}}
& \vqscore{39.02}{\vqgain{+2.93}}
& \vqscore{21.96}{\vqdrop{-7.67}}
& \vqscore{31.38}{\vqgain{+1.34}}
& \vqscore{\textbf{63.76}}{\vqdrop{\textbf{-1.01}}}
& \vqscore{92.26}{\vqdrop{-0.63}}
& \vqscore{65.35}{\vqdrop{-0.50}}
& \vqscore{\textbf{32.17}}{\vqgain{\textbf{+0.07}}}
& \vqscore{\textbf{89.30}}{\vqgain{\textbf{+1.50}}}
& \vqscore{\textbf{68.57}}{\vqdrop{\textbf{-0.11}}} \\
\textbf{+VQ}
& \vqscore{33.94}{\vqgain{+2.67}}
& \vqscore{\textbf{28.66}}{\vqgain{\textbf{+5.49}}}
& \vqscore{36.94}{\vqgain{+0.85}}
& \vqscore{\textbf{31.22}}{\vqgain{\textbf{+1.59}}}
& \vqscore{\textbf{32.69}}{\vqgain{\textbf{+2.65}}}
& \vqscore{63.71}{\vqdrop{-1.06}}
& \vqscore{92.17}{\vqdrop{-0.72}}
& \vqscore{\textbf{65.95}}{\vqgain{\textbf{+0.10}}}
& \vqscore{31.95}{\vqdrop{-0.15}}
& \vqscore{87.50}{\vqdrop{-0.30}}
& \vqscore{68.26}{\vqdrop{-0.42}} \\
\bottomrule
\end{tabular}
\end{adjustbox}

\endgroup
\end{table}

\paragraph{Math:}

As shown in Table~\ref{tab:vq-math-results}, VQ training improves the math average from 30.56 to 34.83 (+4.27). This gain exceeds LoRA (+3.15), although it remains below full training (+6.16). The forgetting behavior reverses this ordering: the general average changes by +0.39 for VQ, compared with -0.42 for LoRA and -0.97 for full training. VQ therefore exceeds LoRA on both reported averages. Relative to full training, it trades 1.89 points of target performance
for a 1.36-point higher general average. The matched comparison also favors \texttt{NCP-ArchPreview}: Full and LoRA reach higher math averages than their OLMo-3 counterparts, while retaining higher
general averages.

\begin{table}[tbp]
\centering
\caption{Mathematics adaptation from the corresponding Stage-1 checkpoints on GSM8K and MATH-500, with general benchmarks
measuring retained capabilities.}
\label{tab:vq-math-results}
\begingroup
\def\vqgain#1{\,{\raisebox{-0.35ex}{\scriptsize\textcolor{green!60!black}{#1}}}}
\def\vqdrop#1{\,{\raisebox{-0.35ex}{\scriptsize\textcolor{red!80!}{#1}}}}
\def\vqscore#1#2{#1#2}
\footnotesize
\setlength{\tabcolsep}{2.6pt}
\renewcommand{\arraystretch}{1.04}

\begin{adjustbox}{max width=\linewidth,center}
\begin{tabular}{@{}lccc@{\hspace{7pt}}cccccc@{}}
\toprule
\multirow{2}{*}{\raisebox{-0.85ex}{\textbf{Model}}} &
\multicolumn{3}{c}{\textbf{Math}} &
\multicolumn{6}{c}{\textbf{General}} \\
\cmidrule(lr){2-4}\cmidrule(lr){5-10}
& GSM8K & MATH-500 & \textbf{Avg.}
& MMLU & ARC-E & HellaSwag & NQ Open & SciQ MC & \textbf{Avg.} \\
\midrule
OLMo-3-7B
& 39.88 & 12.79 & 26.34
& 62.25 & 91.08 & 66.25 & 29.46 & 88.50 & 67.51 \\
+Full
& \vqscore{\textbf{63.23}}{\vqgain{\textbf{+23.35}}}
& \vqscore{9.84}{\vqdrop{-2.95}}
& \vqscore{\textbf{36.54}}{\vqgain{\textbf{+10.20}}}
& \vqscore{60.93}{\vqdrop{-1.32}}
& \vqscore{90.74}{\vqdrop{-0.34}}
& \vqscore{\textbf{65.20}}{\vqdrop{\textbf{-1.05}}}
& \vqscore{28.29}{\vqdrop{-1.17}}
& \vqscore{86.90}{\vqdrop{-1.60}}
& \vqscore{66.41}{\vqdrop{-1.10}} \\
+LoRA
& \vqscore{47.61}{\vqgain{+7.73}}
& \vqscore{\textbf{14.48}}{\vqgain{\textbf{+1.69}}}
& \vqscore{31.05}{\vqgain{+4.71}}
& \vqscore{\textbf{61.21}}{\vqdrop{\textbf{-1.04}}}
& \vqscore{\textbf{91.20}}{\vqgain{\textbf{+0.12}}}
& \vqscore{64.50}{\vqdrop{-1.75}}
& \vqscore{\textbf{29.39}}{\vqdrop{\textbf{-0.07}}}
& \vqscore{\textbf{87.20}}{\vqdrop{\textbf{-1.30}}}
& \vqscore{\textbf{66.70}}{\vqdrop{\textbf{-0.81}}} \\
\midrule
NCP-ArchPreview
& 46.78 & 14.34 & 30.56
& 64.77 & 92.89 & 65.85 & 32.10 & 87.80 & 68.68 \\
+Full
& \vqscore{\textbf{61.03}}{\vqgain{\textbf{+14.25}}}
& \vqscore{12.40}{\vqdrop{-1.94}}
& \vqscore{\textbf{36.72}}{\vqgain{\textbf{+6.16}}}
& \vqscore{63.90}{\vqdrop{-0.87}}
& \vqscore{92.34}{\vqdrop{-0.55}}
& \vqscore{64.90}{\vqdrop{-0.95}}
& \vqscore{29.91}{\vqdrop{-2.19}}
& \vqscore{87.50}{\vqdrop{-0.30}}
& \vqscore{67.71}{\vqdrop{-0.97}} \\
+LoRA
& \vqscore{53.07}{\vqgain{+6.29}}
& \vqscore{14.34}{\vqgain{+0.00}}
& \vqscore{33.71}{\vqgain{+3.15}}
& \vqscore{64.02}{\vqdrop{-0.75}}
& \vqscore{92.09}{\vqdrop{-0.80}}
& \vqscore{65.15}{\vqdrop{-0.70}}
& \vqscore{31.75}{\vqdrop{-0.35}}
& \vqscore{88.30}{\vqgain{+0.50}}
& \vqscore{68.26}{\vqdrop{-0.42}} \\
\textbf{+VQ}
& \vqscore{52.69}{\vqgain{+5.91}}
& \vqscore{\textbf{16.96}}{\vqgain{\textbf{+2.62}}}
& \vqscore{34.83}{\vqgain{+4.27}}
& \vqscore{\textbf{64.11}}{\vqdrop{\textbf{-0.66}}}
& \vqscore{\textbf{92.72}}{\vqdrop{\textbf{-0.17}}}
& \vqscore{\textbf{66.55}}{\vqgain{\textbf{+0.70}}}
& \vqscore{\textbf{32.79}}{\vqgain{\textbf{+0.69}}}
& \vqscore{\textbf{89.20}}{\vqgain{\textbf{+1.40}}}
& \vqscore{\textbf{69.07}}{\vqgain{\textbf{+0.39}}} \\
\bottomrule
\end{tabular}
\end{adjustbox}

\endgroup
\end{table}

\paragraph{Knowledge:}

For knowledge adaptation, Table~\ref{tab:vq-knowledge-results} shows that, for \texttt{NCP-ArchPreview}, VQ training improves TriviaQA exact match from 40.28 to 49.47 (+9.19), while keeping the general average essentially unchanged (+0.03). The target gain is smaller than those of full training (+18.00) and LoRA (+17.48). Relative to OLMo-3, \texttt{NCP-ArchPreview} obtains larger TriviaQA gains with both full training (+18.00 vs. +8.53) and LoRA (+17.48 vs. +8.06). OLMo-3-7B also forgets
more: its general average drops by 1.04 and 1.80 points, respectively, with especially large SciQ MC losses of 4.90 and 9.60 points. Prior work characterizes Transformer feed-forward layers as key-value memories and identifies specific FFN neurons
and mid-layer MLPs that mediate factual recall and support factual editing~\citep{geva-etal-2021-transformer, dai-etal-2022-knowledge,meng2023massediting}. VQ training leaves these backbone MLP parameters fixed and adapts only the concept vocabulary. It therefore cannot directly rewrite factual associations encoded in the backbone, which accounts for its smaller TriviaQA gain relative to full training and LoRA.

\begin{table}[tbp]
\centering
\caption{Results after knowledge adaptation, initialized from the
corresponding Stage-1 checkpoints.}
\label{tab:vq-knowledge-results}
\begingroup
\def\vqgain#1{\,{\raisebox{-0.35ex}{\scriptsize\textcolor{green!60!black}{#1}}}}
\def\vqdrop#1{\,{\raisebox{-0.35ex}{\scriptsize\textcolor{red!80!}{#1}}}}
\def\vqscore#1#2{#1#2}
\footnotesize
\setlength{\tabcolsep}{2.6pt}
\renewcommand{\arraystretch}{1.04}

\begin{tabular}{@{}lc@{\hspace{7pt}}cccccc@{}}
\toprule
\multirow{2}{*}{\raisebox{-0.85ex}{\textbf{Model}}} &
\multicolumn{1}{c}{\textbf{Knowledge}} &
\multicolumn{6}{c}{\textbf{General}} \\
\cmidrule(lr){2-2}\cmidrule(lr){3-8}
& TriviaQA EM
& MMLU & ARC-E & HellaSwag & NQ Open & SciQ MC & \textbf{Avg.} \\
\midrule
OLMo-3-7B
& 48.49
& 62.25 & 91.08 & 66.25 & 29.46 & 88.50 & 67.51 \\
+Full
& \vqscore{\textbf{57.02}}{\vqgain{\textbf{+8.53}}}
& \vqscore{61.73}{\vqdrop{-0.52}}
& \vqscore{90.57}{\vqdrop{-0.51}}
& \vqscore{66.65}{\vqgain{+0.40}}
& \vqscore{\textbf{29.80}}{\vqgain{\textbf{+0.34}}}
& \vqscore{\textbf{83.60}}{\vqdrop{\textbf{-4.90}}}
& \vqscore{\textbf{66.47}}{\vqdrop{\textbf{-1.04}}} \\
+LoRA
& \vqscore{56.55}{\vqgain{+8.06}}
& \vqscore{\textbf{62.19}}{\vqdrop{\textbf{-0.06}}}
& \vqscore{\textbf{90.82}}{\vqdrop{\textbf{-0.26}}}
& \vqscore{\textbf{67.50}}{\vqgain{\textbf{+1.25}}}
& \vqscore{29.13}{\vqdrop{-0.33}}
& \vqscore{78.90}{\vqdrop{-9.60}}
& \vqscore{65.71}{\vqdrop{-1.80}} \\
\midrule
NCP-ArchPreview
& 40.28
& 64.77 & 92.89 & 65.85 & 32.10 & 87.80 & 68.68 \\
+Full
& \vqscore{\textbf{58.28}}{\vqgain{\textbf{+18.00}}}
& \vqscore{\textbf{64.70}}{\vqdrop{\textbf{-0.07}}}
& \vqscore{\textbf{92.76}}{\vqdrop{\textbf{-0.13}}}
& \vqscore{67.05}{\vqgain{+1.20}}
& \vqscore{30.62}{\vqdrop{-1.48}}
& \vqscore{86.80}{\vqdrop{-1.00}}
& \vqscore{68.39}{\vqdrop{-0.29}} \\
+LoRA
& \vqscore{57.76}{\vqgain{+17.48}}
& \vqscore{64.07}{\vqdrop{-0.70}}
& \vqscore{92.51}{\vqdrop{-0.38}}
& \vqscore{\textbf{67.10}}{\vqgain{\textbf{+1.25}}}
& \vqscore{31.14}{\vqdrop{-0.96}}
& \vqscore{87.90}{\vqgain{+0.10}}
& \vqscore{68.54}{\vqdrop{-0.14}} \\
\textbf{+VQ}
& \vqscore{49.47}{\vqgain{+9.19}}
& \vqscore{64.37}{\vqdrop{-0.40}}
& \vqscore{92.38}{\vqdrop{-0.51}}
& \vqscore{66.65}{\vqgain{+0.80}}
& \vqscore{\textbf{31.63}}{\vqdrop{\textbf{-0.47}}}
& \vqscore{\textbf{88.50}}{\vqgain{\textbf{+0.70}}}
& \vqscore{\textbf{68.71}}{\vqgain{\textbf{+0.03}}} \\
\bottomrule
\end{tabular}

\endgroup
\end{table}

\begin{figure}[tbp]
    \centering
    \includegraphics[width=\linewidth]{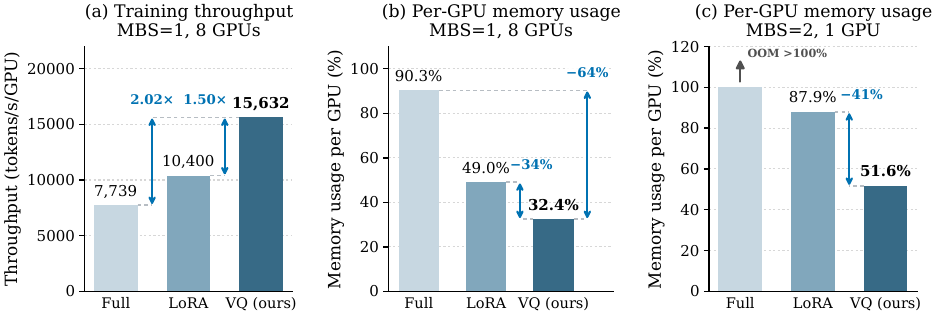}
    \caption{Training speed and per-GPU memory usage for full training, LoRA, and VQ training. Speed is measured in tokens/s/GPU, memory is reported as a percentage of capacity, and MBS denotes micro-batch size.}
    \label{fig:vq-adaptation-efficiency}
\end{figure}

\paragraph{Training Throughput and Memory Usage:}
Figure~\ref{fig:vq-adaptation-efficiency} compares the training throughput and GPU memory usage of VQ, LoRA, and full training. With a micro-batch size of 1 on 8 GPUs, VQ reaches 15,632 tokens/s/GPU, compared with 10,400 for LoRA and 7,739 for full training, corresponding to 1.50$\times$ and 2.02$\times$ higher throughput, respectively. Although VQ and LoRA both optimize 17M parameters, VQ achieves 50\% higher throughput, while full training updates all 8.9B
parameters. VQ also has the lowest memory footprint, using 32.4\% of the available memory, compared with 49.0\% for LoRA and 90.3\% for full training. With a micro-batch size of 2 on a single GPU, VQ uses 51.6\% of memory, while LoRA reaches 87.9\% and full training exceeds memory capacity. Overall, VQ combines parameter efficiency relative to full training with higher throughput and lower memory usage than both baselines.

\subsection{Multi-Token Prediction on NCP-ArchPreview}
\label{sec:hlm-mtp-interaction}

\begin{figure}[tbp]
    \centering
    \includegraphics[width=\linewidth]{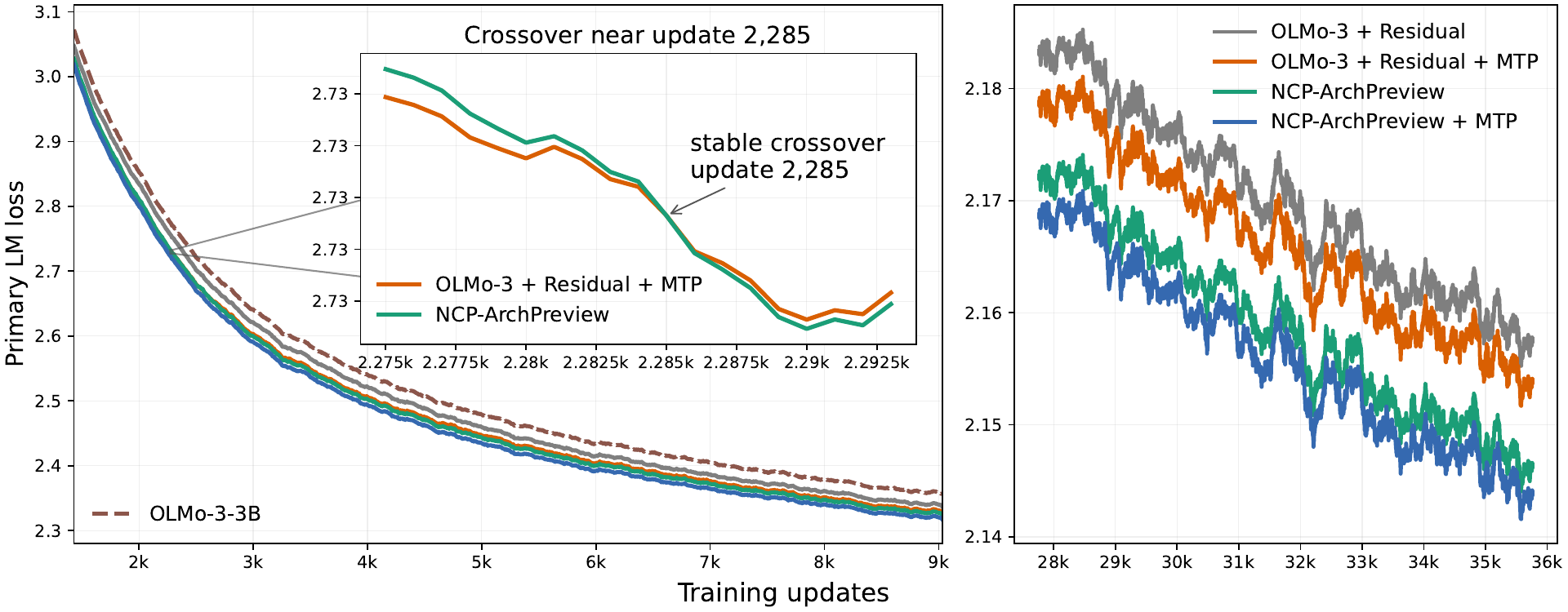}
    \caption{NTP loss over the first 150B training tokens, smoothed with a 200-update exponential moving average. The left panel includes the crossover in early training. The right panel shows the final 8k updates of the compared variants.}
    \label{fig:hlm-mtp-loss-trajectories}
\end{figure}

We build \texttt{NCP-ArchPreview} and its baseline from OLMo-3-3B. The 3B-scale \texttt{NCP-ArchPreview} organizes the model into an 8-layer Token Encoder, a 4-layer Concept Module, and a 7-layer Token Decoder. For a fair comparison, we further modify OLMo-3-3B by dividing its 16 Transformer layers into groups of 8/1/7 layers and placing the same hierarchical residual connections (Section~\ref{subsec:hierarchical-residual}) at corresponding depths, which is denoted as OLMo-3-3B + Residual.
For each architecture, we train a variant with one additional MTP layer.
All four controlled variants are trained on the first 150B tokens of the
pretraining corpus.
Following DeepSeek-V3~\citep{deepseekai2024deepseekv3}, the main model retains
the next-token prediction (NTP) objective, while the added MTP layer predicts
the token two positions ahead with loss weight $0.3$.

\begin{table}[t]
  \centering
  \caption{Comparison of residual-aligned OLMo-3-3B variants and the
  3B-scale NCP-ArchPreview model. LM loss is averaged over the first 150B training tokens and excludes auxiliary losses.}
  \label{tab:hlm-mtp-interaction}
  \small
  \setlength{\tabcolsep}{4pt}
  \renewcommand{\arraystretch}{1.12}
  \begin{tabular}{@{}p{0.28\linewidth}cccc@{}}
      \toprule
      Model & \shortstack{Avg. LM\\loss}
      & \shortstack{$\Delta$ loss\\vs. OLMo-3\\+ Residual}
      & \shortstack{Training\\FLOPs}
      & \shortstack{$\Delta$ FLOPs\\vs. OLMo-3\\+ Residual} \\
      \midrule
      OLMo-3-3B + Residual
      & 2.3805 & 0.0000 & $3.245 \times 10^{21}$ & $0.00\%$ \\
      OLMo-3-3B + Residual + MTP
      & 2.3739 & $-0.0065$ & $3.750 \times 10^{21}$ & $+15.55\%$ \\
      \midrule
      NCP-ArchPreview
      & 2.3707 & $-0.0098$ & $3.227 \times 10^{21}$ & $-0.57\%$ \\
      NCP-ArchPreview + MTP
      & 2.3664 & $-0.0141$ & $3.731 \times 10^{21}$ & $+14.98\%$ \\
      \bottomrule
  \end{tabular}
\end{table}

Figure~\ref{fig:hlm-mtp-loss-trajectories} shows how the ordering develops over
training. OLMo-3-3B + Residual with MTP initially has lower loss than NCP-ArchPreview, but \texttt{NCP-ArchPreview} crosses below it near update 2.285k. Over the final 8k updates, the curves consistently maintain the ordering \texttt{NCP-ArchPreview} with MTP, \texttt{NCP-ArchPreview}, OLMo-3-3B + Residual with MTP, and OLMo-3-3B + Residual, from lowest to highest loss. \texttt{NCP-ArchPreview} without MTP eventually outperforms OLMo-3-3B + Residual with MTP as training progresses, while adding MTP to \texttt{NCP-ArchPreview} provides a further improvement.

Table~\ref{tab:hlm-mtp-interaction} summarizes the complete training interval over the first 150B tokens. \texttt{NCP-ArchPreview} improves over OLMo-3-3B + Residual by 0.0098, which is greater than the 0.0065 improvement from adding MTP to OLMo-3-3B + Residual. \texttt{NCP-ArchPreview} with MTP is 0.0075 lower than OLMo-3-3B + Residual with MTP while using slightly fewer training FLOPs.

\subsection{Predicted Concept Improves Block-Parallel Speculative Drafting}
\label{sec:hlm-speculative-drafting}

Recent advances in block-diffusion language modeling have demonstrated the feasibility of generating multiple tokens in parallel within each block~\citep{arriola2025blockdiffusion,cheng2026sdar}. Building on this capability, block-parallel draft models have emerged as a promising direction for speculative decoding~\citep{chen2026dflash,cheng2026dspark,inco2026dflash2}. These models reduce drafting
latency by proposing an entire token block in one forward pass, but their efficiency remains limited by the difficulty of maintaining a coherent prediction trajectory across multiple future positions.

The results in Section~\ref{sec:hlm-mtp-interaction} suggest that concept representations contain information useful for multi-token prediction. We therefore investigate whether the states produced by the Concept Module can improve the acceptance length of a block-parallel Drafter.

\paragraph{Experimental Design:}

Our baseline combines complementary architectural improvements from DFlash2 and DFlare. Following DFlash2~\citep{inco2026dflash2}, we use a two-tap dynamic convolution to model local dependencies between adjacent draft positions and a lightweight path selector to select a coherent sequence from the top-16 candidates at each position. Following DFlare~\citep{zhang2026dflare}, each Drafter layer learns its own weighted combination of five Target hidden states, providing layer-specific access to Target representations at different semantic depths.

Our \textit{Baseline+Concept} variant keeps the baseline unchanged and adds a causal hierarchical signal. For a verified prefix ending at position $t$, we use the concept representation $c_{\lfloor t/k \rfloor}$ from the last fully completed Target chunk. At Drafter layer $\ell$, the concept representation is incorporated into every proposal position as
\begin{equation}
    \widetilde{s}^{(\ell)}_{t,j}
    =
    s^{(\ell)}_{t,j}
    +
    \tanh\!\left(g^{(\ell)}\right)
    \odot
    \operatorname{RMSNorm}_{\ell}(c_{\lfloor t/k \rfloor}),
    \qquad j=1,\ldots,K.
\end{equation}
The gates are initialized to zero, allowing the Drafter to gradually incorporate the concept signal. This adds only 0.04M parameters to the 1.1B-parameter Drafter, without introducing additional Target-model computation.

\paragraph{Experimental setting:}

The \textit{Baseline} and \textit{Baseline+Concept} Drafters are trained using the same online Target distillation pipeline, data order, random seed, and training budget. Both models use a sequence length of 8192, 512 training anchors per sequence, a global batch size of 512, and a fixed 5B-token training subset repeated for 10 epochs. We evaluate the Drafter on GSM8K, MATH, HumanEval and MBPP using exact speculative verification with a proposal horizon of 16 draft tokens.

Following DSpark, we report mean accepted length (MAL), defined as
\begin{equation}
    \operatorname{MAL}
    =
    \frac{1}{R}\sum_{r=1}^{R} c_r,
\end{equation}
where $c_r$ is the number of tokens committed in verification round $r$, including the Target correction or bonus token when present.

\begin{table}[tbp]
    \centering
    \small
    \setlength{\tabcolsep}{6pt}
    \caption{Mean accepted length on the evaluation set.}
    \label{tab:hlm-speculative-drafting}
    \begin{tabular}{lccc}
        \toprule
        Benchmark & Baseline & Baseline+Concept & Relative gain \\
        \midrule
        GSM8K     & 6.351 & \textbf{6.537} & $+2.93\%$ \\
        MATH      & 6.105 & \textbf{6.240} & $+2.22\%$ \\
        HumanEval & 5.432 & \textbf{5.845} & $+7.59\%$ \\
        MBPP      & 5.844 & \textbf{6.099} & $+4.37\%$ \\
        \midrule
        \textbf{Macro average}
        & 5.933 & \textbf{6.180} & $\mathbf{+4.17\%}$ \\
        \bottomrule
    \end{tabular}
\end{table}

\paragraph{Results:}

As shown in Table~\ref{tab:hlm-speculative-drafting}, concept conditioning improves MAL consistently across all four benchmarks. The macro-average MAL increases from 5.933 to 6.180, corresponding to a $4.17\%$ relative improvement. The largest gain is observed on HumanEval, where MAL improves by $7.59\%$. These results indicate that the chunk-level concept representation complements token-level Target features with broader predictive context, improving the consistency of block-parallel proposals with negligible parameter overhead.

\section{Related Work}
\label{sec:related}

\subsection{Abstract-Level Prediction}

A growing line of research predicts representations rather than reconstructing fine-grained inputs. Joint-embedding predictive architectures learn representations by predicting latent targets that retain semantic structure while discarding input details that are difficult to predict and unnecessary for the learned
representation~\citep{assran2023selfsupervisedlearningimagesjointembedding}. Large Concept Models apply abstract-level autoregressive modeling to language by mapping complete sentences into a shared continuous embedding space~\citep{lcmteam2024largeconceptmodelslanguage}. These approaches motivate targets above raw input units, but use continuous prediction spaces that are typically defined by a separate encoder. Concept-level language modeling instead learns a higher-level prediction space within the language model while retaining token-level generation.

\subsection{Hierarchical and Latent-Space Language Modeling}

Hierarchical language models introduce a shorter high-level sequence over
groups of fine-grained units. The Hourglass Transformer compresses and later upsamples token sequences~\citep{hourglass}, while MegaByte performs global modeling over fixed byte patches and local modeling within each
patch~\citep{yu2023megabytepredictingmillionbytesequences}. BLT and H-Net replace fixed tokenization with data-dependent byte patches~\citep{pagnoni2024bytelatenttransformerpatches,
hwang2025dynamicchunkingendtoendhierarchical}. At the token level, ContextLM learns predictive context embeddings, and DLCM constructs higher-level units through dynamic chunking~\citep{dai2025contextlevellanguagemodelinglearning,
qu2026dynamiclargeconceptmodels}. These methods primarily alter the
computational granularity or the mechanism used to construct latent units.

Other work changes the prediction objective or introduces latent states into the generative process. Multi-token prediction adds auxiliary heads for several future tokens while retaining token-level targets~\citep{gloeckle2024betterfasterlarge}. Methods based on latent reasoning or continuous concepts instead interleave or predict continuous hidden
states~\citep{su2025tokenassortedmixinglatent,
hao2025traininglargelanguagemodels}.
ConceptLM is the direct foundation of our work: it introduced NCP as a
discrete concept-level objective, learned a product-quantized concept
vocabulary jointly with the language model, and conditioned token generation on predicted concepts~\citep{liu2026conceptpredictiondiscretelatent}. ConceptLM evaluated models trained from scratch up to 1.5B parameters and added NCP to an existing 8B model through continual pre-training. \texttt{NCP-ArchPreview} instead studies an 8.9B model derived from the Olmo-3-7B architecture in which NCP is active from the start of pre-training over several
trillion tokens and during subsequent training stages.

\subsection{Residual Connections in Deep Transformers}

Standard residual connections accumulate cross-layer information into a
single additive stream, motivating mechanisms that selectively reuse earlier representations. DenseFormer uses input-independent learned weights to average previous block outputs across depth~\citep{pagliardini2024denseformerenhancinginformationflow}.
DeepCrossAttention uses input-dependent weights to combine layer outputs and additionally introduces depth-wise cross-attention~\citep{pmlr-v267-heddes25a}.
In parallel, MUDDFormer predicts token-conditioned dense-connection weights separately for the query, key, value, and residual streams~\citep{xiao2025muddformer}. Attention
Residuals instead applies softmax attention over preceding layer outputs, with a blockwise variant for scalable training~\citep{kimiteam2026attentionresiduals}. Depth-Attention places cross-layer selection inside self-attention by mixing earlier value states through the existing QKV pathway~\citep {zeng2026depthattentioncrosslayervaluemixing}. \texttt{NCP-ArchPreview} adopts MUDDFormer's single-stream DD formulation as Intra-Module Residual Connections (IRC), enabling full-history reuse within each
module. Cross-Module Residual Connections (CRC) further enable target-conditioned depth selection across the Encoder, Concept Module, and Decoder.

\section{Limitations}
\label{sec:limitations}

\paragraph{Extension to long-context training:}
This report focuses on \texttt{NCP-ArchPreview} after 5.73T tokens of pre-training and
100B tokens of mid-training at standard context lengths. Long-context training
is not included in the current architecture preview and remains part of our
planned full training recipe. Since the concept-level pathway operates on a
compressed sequence, longer contexts may provide a particularly favorable
setting for latent-space modeling. Extending \texttt{NCP-ArchPreview} to this regime
will allow us to study how concept prediction and hierarchical residual routing
scale to substantially longer dependencies.

\paragraph{Relationship between language-modeling loss and downstream performance:}
\texttt{NCP-ArchPreview} maintains a clear token-level language-modeling loss advantage
over OLMo-3-7B throughout both pre-training and mid-training, together with
positive aggregate downstream gains at both checkpoints. The downstream
improvement is more pronounced after pre-training, where the overall average
increases by 2.45 points, than after mid-training, where the gain is 0.59
points and varies across benchmarks. Similar stage- and task-dependent
relationships between cross-entropy loss and downstream capability have also
been observed in recent work~\citep{patel2026forecastingdownstreamperformancellms,yano2026pretrainingllmwithoutlearningratedecay}.

These results establish a consistent optimization advantage for
\texttt{NCP-ArchPreview} while suggesting that the conversion from lower
language-modeling loss to downstream performance depends on the training
distribution and evaluation domain. Future work will explore additional
mid-training recipes, capability-aware data selection, and checkpoint
selection criteria that combine language-modeling loss with downstream and
proxy evaluations.
\section{Conclusion}
\label{sec:conclusion}

This work establishes that latent representations learned by a language model can serve as first-class prediction targets at trillion-token scale. \texttt{NCP-ArchPreview} turns this idea into an 8.9B-parameter architecture that jointly models tokens and learned discrete concepts while preserving standard autoregressive generation. After pre-training on 5.73T tokens, it reaches the final loss of OLMo-3-7B with only 51.3\% of its training tokens and improves the downstream macro-average by 2.45 points, including a 5.99-point gain on GSM8K. Controlled experiments show complementary gains from the latent architecture, hierarchical routing, and NCP; the complete model approaches a parameter-aligned 40-layer Transformer with only 85\% of its computation, while scaling-law experiments indicate a $1.74\times$ improvement in compute efficiency. The learned concept space also supports efficient domain adaptation through only 17M VQ parameters and improves the mean accepted length of a DFlash2-based speculative drafter by 4.17\% with negligible overhead. Together, these results provide the largest-scale validation of latent-space language modeling to date and position joint token and concept prediction as a practical, scalable blueprint for more capable and efficient language models. Extending this architecture through long-context training and translating its optimization advantages more consistently into downstream capability remain important directions for future work.

\IfFileExists{references.bib}{\bibliography{references}}{}

\clearpage
\appendix
\section{Author List}
\label{app:authors}

\subsection{Full List (Alphabetically)}

\noindent

Jiaqi Cao; Chiyu Chen; Shuang Cheng; Xu Cheng; Beiya Dai; Yufan Feng; Kewen Ge; Ruijun Ge; Jiayi Huang; Yang Jiao; Dahua Lin; Zhouhan Lin; Yifan Liu; Yuliang Liu; Biqing Qi; Mowen Ruan; Junzhe Shen; Yunchong Song; Hao Sun; Zhongbo Tian; Yixuan Wang; Rubin Wei; Jiaxin Xiong; Kangyu Yang; Qian Yao; Qi Zhang; Bowen Zhou

\subsection{Contributions}

\begin{tabular}{@{}p{0.28\linewidth}@{\hspace{0.02\linewidth}}p{0.7\linewidth}@{}}

\textbf{Architecture Design}
& Zhouhan Lin; Yuliang Liu \\

\textbf{Architecture Optimization}
& Yuliang Liu; Jiayi Huang; Jiaxin Xiong; Beiya Dai \\

\textbf{Model Training}
& Yuliang Liu; Jiayi Huang; Yang Jiao; Junzhe Shen; Rubin Wei; Xu Cheng; Yixuan Wang \\

\textbf{Scaling Ladder}
& Mowen Ruan; Yuliang Liu \\

\textbf{Ablation and Analysis}
& Jiayi Huang; Rubin Wei; Yufan Feng; Yuliang Liu \\

\textbf{Inference and Deployment}
& Yixuan Wang; Shuang Cheng; Biqing Qi; Qian Yao; Kewen Ge \\

\textbf{Evaluation}
& Yixuan Wang; Yuliang Liu \\

\textbf{Data Analysis}
& Hao Sun; Yunchong Song; Yifan Liu; Yuliang Liu \\

\textbf{Infrastructure}
& Ruijun Ge; Chiyu Chen; Zhongbo Tian \\

\textbf{Project Leader}
& Zhouhan Lin \\

\textbf{Project Advisors}
& Zhouhan Lin; Qi Zhang; Dahua Lin; Bowen Zhou

\end{tabular}

\par\vspace{0.6em}
{\small
\noindent
\faEnvelope\enspace
{Correspondence to:}\enspace
\href{mailto:liuyl03181@gmail.com}{liuyl03181@gmail.com}
\enspace\textbar\enspace
\href{mailto:hantek@sjtu.edu.cn}{hantek@sjtu.edu.cn}
}

\section{Additional Architecture Details}
\label{app:architecture}

\subsection{Complete Architecture Configuration}
\label{app:architecture-configuration}

Table~\ref{tab:ncp-architecture} summarizes the complete architecture
configuration of \texttt{NCP-ArchPreview}.

\begin{table}[tbp]
\centering
\caption{Architecture configuration of \texttt{NCP-ArchPreview}.}
\label{tab:ncp-architecture}
\small
\begin{tabularx}{\linewidth}{@{}l>{\raggedright\arraybackslash}X@{}}
\toprule
Component & Configuration \\
\midrule
Model class & Latent-space autoregressive language model \\
Total parameters & 8.94B \\
Token Encoder & 16 causal Transformer layers \\
Concept Module & 8 causal Transformer layers \\
Token Decoder & 16 causal Transformer layers \\
Hidden dimension & $4{,}096$ \\
FFN dimension & $11{,}008$ \\
Attention heads & 32 \\
Key--value groups & 32 \\
Attention-head dimension & 128 \\
Vocabulary size & $100{,}278$ \\
Maximum training length & $8{,}192$ tokens \\
Position encoding & RoPE with base $500{,}000$ \\
Token-level local-attention window & $4{,}096$ tokens \\
Full-attention pattern & Every fourth layer uses full attention \\
Activation & SwiGLU \\
Normalization & RMSNorm with $\epsilon=10^{-6}$ \\
Attention normalization & Layer-wise QK RMSNorm \\
Attention and hidden dropout & 0 \\
Concept compression factor $k$ & 4 token states per concept \\
Concept sequence length & $M=T/k$ after padding and masking \\
PQ segments $S$ & 32 \\
Codewords per segment $N$ & 128 \\
Codeword dimension & 128 \\
Latent vocabulary capacity & $N^S=128^{32}$ possible code combinations \\
Token-level objective & Next-token prediction (NTP) \\
Latent-space objective & Next-concept prediction (NCP) \\
Concept predictor & Dedicated Concept Module \\
Concept feedback & Causally shifted, repeated $k$ times, and added to token states \\
Intra-module residual routing & IRC in the Token Encoder, Concept Module, and Token Decoder \\
Cross-module residual routing &
Token Encoder$\rightarrow$Concept Module,
Token Encoder$\rightarrow$Token Decoder, and
Concept Module$\rightarrow$Token Decoder \\
Parameter precision & BF16 \\
\bottomrule
\end{tabularx}
\end{table}

\subsection{Forward-Pass Pseudocode}
\label{app:forward-pseudocode}

Algorithm~\ref{alg:ncp-forward} summarizes the end-to-end training
forward pass. The intra-module residual connections (IRC) and
cross-module residual connections (CRC) follow
Eqs.~\ref{eq:irc-block-output}--\ref{eq:crc-update} in
Section~\ref{subsec:hierarchical-residual}.
The causal alignment between concept-level and token-level states follows
Eq.~\ref{eq:causal-concept-broadcast}.

For compactness, $\mathcal{S}^{e}$ and $\mathcal{S}^{c}$ denote the
intermediate layer states exported by the Token Encoder and Concept Module,
respectively. The operator
$\operatorname{ChunkPool}_{k}$ converts token-level residual states to
concept resolution, whereas
$\operatorname{CausalShiftAndRepeat}$ converts concept-level states back
to token resolution without exposing future concepts.

\providecommand{\ConceptStageOneModel}{\texttt{NCP\_\allowbreak ArchPreview\_\allowbreak dolma3\_8.9B\_\allowbreak Stage1\_V1}}
\providecommand{\ConceptStageTwoVOneModel}{\texttt{NCP\_\allowbreak ArchPreview\_\allowbreak dolma3\_8.9B\_\allowbreak Stage2\_V1}}
\providecommand{\ConceptStageTwoVTwoModel}{\texttt{NCP\_\allowbreak ArchPreview\_\allowbreak dolma3\_8.9B\_\allowbreak Stage2\_V2}}
\providecommand{\ConceptStageTwoVThreeModel}{\texttt{NCP\_\allowbreak ArchPreview\_\allowbreak dolma3\_8.9B\_\allowbreak Stage2\_V3}}

\providecommand{\ProxyCodeStageOne}{0.9338}
\providecommand{\ProxyCodeStageTwoVOne}{0.7330}
\providecommand{\ProxyCodeStageTwoVTwo}{0.7501}
\providecommand{\ProxyCodeStageTwoVThree}{0.7598}
\providecommand{\ProxyCodeStageOneToStageTwoVOneReduction}{21.5\%}
\providecommand{\ProxyCodeStageTwoVOneVersusVThreeReduction}{3.5\%}
\providecommand{\ProxyCompetitionMathStageOne}{0.7979}
\providecommand{\ProxyCompetitionMathStageTwoVOne}{0.5742}
\providecommand{\ProxyCompetitionMathStageTwoVTwo}{0.5949}
\providecommand{\ProxyCompetitionMathStageTwoVThree}{0.6075}
\providecommand{\ProxyCompetitionMathStageOneToStageTwoVOneReduction}{28.0\%}
\providecommand{\ProxyCompetitionMathStageTwoVOneVersusVThreeReduction}{5.5\%}
\providecommand{\ProxyAcademicSTEMStageOne}{1.0570}
\providecommand{\ProxyAcademicSTEMStageTwoVOne}{0.8909}
\providecommand{\ProxyAcademicSTEMStageTwoVTwo}{0.9007}
\providecommand{\ProxyAcademicSTEMStageTwoVThree}{0.9058}
\providecommand{\ProxyAcademicSTEMStageOneToStageTwoVOneReduction}{15.7\%}
\providecommand{\ProxyAcademicSTEMStageTwoVOneVersusVThreeReduction}{1.6\%}
\providecommand{\ProxyLogicalReasoningStageOne}{0.9692}
\providecommand{\ProxyLogicalReasoningStageTwoVOne}{0.8148}
\providecommand{\ProxyLogicalReasoningStageTwoVTwo}{0.8314}
\providecommand{\ProxyLogicalReasoningStageTwoVThree}{0.8355}
\providecommand{\ProxyLogicalReasoningStageOneToStageTwoVOneReduction}{15.9\%}
\providecommand{\ProxyLogicalReasoningStageTwoVOneVersusVThreeReduction}{2.5\%}
\providecommand{\ProxyHellaMarginStageOne}{0.09610}
\providecommand{\ProxyHellaMarginStageTwoVOne}{0.09678}
\providecommand{\ProxyHellaMarginStageTwoVTwo}{0.09932}
\providecommand{\ProxyHellaMarginStageTwoVThree}{0.10083}
\providecommand{\ProxyHellaVThreeMinusVTwoMargin}{0.00151}
\providecommand{\ProxyHellaVThreeMinusVTwoCILow}{0.00097}
\providecommand{\ProxyHellaVThreeMinusVTwoCIHigh}{0.00205}
\providecommand{\CoreHellaVTwoMinusVThree}{0.05}
\providecommand{\ProxyAlignmentRhoOne}{1.000}
\providecommand{\ProxyAlignmentPairsOne}{6/6}
\providecommand{\ProxyAlignmentStageTwoRhoOne}{1.000}
\providecommand{\ProxyAlignmentStageTwoPairsOne}{3/3}
\providecommand{\ProxyAlignmentRhoTwo}{1.000}
\providecommand{\ProxyAlignmentPairsTwo}{6/6}
\providecommand{\ProxyAlignmentStageTwoRhoTwo}{1.000}
\providecommand{\ProxyAlignmentStageTwoPairsTwo}{3/3}
\providecommand{\ProxyAlignmentRhoThree}{1.000}
\providecommand{\ProxyAlignmentPairsThree}{6/6}
\providecommand{\ProxyAlignmentStageTwoRhoThree}{1.000}
\providecommand{\ProxyAlignmentStageTwoPairsThree}{3/3}
\providecommand{\ProxyAlignmentRhoFour}{1.000}
\providecommand{\ProxyAlignmentPairsFour}{6/6}
\providecommand{\ProxyAlignmentStageTwoRhoFour}{1.000}
\providecommand{\ProxyAlignmentStageTwoPairsFour}{3/3}
\providecommand{\ProxyAlignmentRhoFive}{1.000}
\providecommand{\ProxyAlignmentPairsFive}{6/6}
\providecommand{\ProxyAlignmentStageTwoRhoFive}{1.000}
\providecommand{\ProxyAlignmentStageTwoPairsFive}{3/3}
\providecommand{\ProxyAlignmentRhoSix}{1.000}
\providecommand{\ProxyAlignmentPairsSix}{6/6}
\providecommand{\ProxyAlignmentStageTwoRhoSix}{1.000}
\providecommand{\ProxyAlignmentStageTwoPairsSix}{3/3}
\providecommand{\ProxyAlignmentRhoSeven}{0.200}
\providecommand{\ProxyAlignmentPairsSeven}{3/6}
\providecommand{\ProxyAlignmentStageTwoRhoSeven}{-1.000}
\providecommand{\ProxyAlignmentStageTwoPairsSeven}{0/3}
\providecommand{\ProxyAlignmentRhoEight}{0.800}
\providecommand{\ProxyAlignmentPairsEight}{5/6}
\providecommand{\ProxyAlignmentStageTwoRhoEight}{0.500}
\providecommand{\ProxyAlignmentStageTwoPairsEight}{2/3}
\providecommand{\ProxyFitStageTwoRSquaredHumanEval}{0.999}
\providecommand{\ProxyFitAllFourRSquaredHumanEval}{0.928}
\providecommand{\ProxyFitStageTwoRSquaredMBPP}{0.906}
\providecommand{\ProxyFitAllFourRSquaredMBPP}{0.974}
\providecommand{\ProxyFitStageTwoRSquaredMATHFiveHundred}{0.885}
\providecommand{\ProxyFitAllFourRSquaredMATHFiveHundred}{0.992}
\providecommand{\ProxyFitStageTwoRSquaredMinerva}{0.892}
\providecommand{\ProxyFitAllFourRSquaredMinerva}{0.995}
\providecommand{\ProxyFitStageTwoRSquaredMMLUSTEM}{0.962}
\providecommand{\ProxyFitAllFourRSquaredMMLUSTEM}{0.651}
\providecommand{\ProxyFitStageTwoRSquaredBBH}{0.531}
\providecommand{\ProxyFitAllFourRSquaredBBH}{0.968}
\providecommand{\DownstreamHumanEvalStageOne}{31.27}
\providecommand{\DownstreamHumanEvalStageTwoVOne}{45.60}
\providecommand{\DownstreamHumanEvalStageTwoVTwo}{42.19}
\providecommand{\DownstreamHumanEvalStageTwoVThree}{39.96}
\providecommand{\DownstreamMBPPStageOne}{36.09}
\providecommand{\DownstreamMBPPStageTwoVOne}{50.91}
\providecommand{\DownstreamMBPPStageTwoVTwo}{49.34}
\providecommand{\DownstreamMBPPStageTwoVThree}{46.60}
\providecommand{\DownstreamMATHFiveHundredStageOne}{14.34}
\providecommand{\DownstreamMATHFiveHundredStageTwoVOne}{43.74}
\providecommand{\DownstreamMATHFiveHundredStageTwoVTwo}{41.66}
\providecommand{\DownstreamMATHFiveHundredStageTwoVThree}{37.21}
\providecommand{\DownstreamMinervaStageOne}{15.13}
\providecommand{\DownstreamMinervaStageTwoVOne}{42.20}
\providecommand{\DownstreamMinervaStageTwoVTwo}{40.39}
\providecommand{\DownstreamMinervaStageTwoVThree}{36.70}
\providecommand{\DownstreamMMLUSTEMStageOne}{55.88}
\providecommand{\DownstreamMMLUSTEMStageTwoVOne}{61.84}
\providecommand{\DownstreamMMLUSTEMStageTwoVTwo}{59.85}
\providecommand{\DownstreamMMLUSTEMStageTwoVThree}{57.72}
\providecommand{\DownstreamBBHStageOne}{52.41}
\providecommand{\DownstreamBBHStageTwoVOne}{63.23}
\providecommand{\DownstreamBBHStageTwoVTwo}{62.90}
\providecommand{\DownstreamBBHStageTwoVThree}{60.44}
\providecommand{\DownstreamHellaSwagStageOne}{65.85}
\providecommand{\DownstreamHellaSwagStageTwoVOne}{66.40}
\providecommand{\DownstreamHellaSwagStageTwoVTwo}{67.30}
\providecommand{\DownstreamHellaSwagStageTwoVThree}{67.25}
\providecommand{\DownstreamHumanEvalVOneMinusVThree}{5.64}
\providecommand{\DownstreamMBPPVOneMinusVThree}{4.31}
\providecommand{\DownstreamMATHFiveHundredVOneMinusVThree}{6.52}
\providecommand{\DownstreamMinervaVOneMinusVThree}{5.50}
\providecommand{\DownstreamMMLUSTEMVOneMinusVThree}{4.11}
\providecommand{\DownstreamBBHVOneMinusVThree}{2.79}

\section{Analysis of OLMo-3 Mid-Training Data with Proxy Metrics}
\label{sec:mid-training-data-proxy}

The preceding sections describe the model architecture and training setup. We
next examine whether candidate data recipes for the second training stage can
be compared without repeatedly executing the full downstream evaluation suite.
To this end, we evaluate three matched-budget Stage-2 recipes with a fixed,
capability-specific held-out set. The Stage-1 checkpoint is included as a
cross-stage reference, while recipe selection is based on comparisons among
the three Stage-2 variants.

\subsection{Held-out Evaluation and Scoring}
\label{sec:proxy-held-out-scoring}

The expanded held-out set is constructed from 177,202 expert trajectories spanning 63 sources and six domains, corresponding to approximately 100M teacher-suffix tokens. All trajectories are produced by a single frozen Qwen3.7-Plus teacher under a common generation protocol. The prompt, expert suffix, source identity, and capability annotation are fixed before candidate checkpoints are evaluated, ensuring that every recipe is compared on an identical scoring view. Invalid generations are excluded, and a more restrictive trusted subset is retained for sensitivity analysis.

\begin{algorithm}[H]
\caption{End-to-end training of \texttt{NCP-ArchPreview} with
hierarchical residual routing}
\label{alg:ncp-forward}
\begin{algorithmic}[1]
\Require Tokens $x_{1:T}$; compression factor $k=4$; 
PQ codebooks
$\mathcal{E}^{s}=\{\mathbf{e}_{n}^{s}\}_{n=1}^{N}$,
$s=1,\ldots,S$
\Require Loss weights $\alpha$ and $\beta$

\State $M\gets \lfloor T/k \rfloor$

\State $(\mathbf{H},\mathcal{S}^{e})
\gets
\operatorname{TokenEncoder}_{\theta_e}^{\mathrm{IRC}}(x_{1:T})$
\Comment{IRC inside the TokenEncoder}

\State $\mathbf{C}
\gets
\operatorname{MeanPool}_{k}(\mathbf{H})$

\State $\overline{\mathcal{S}}^{e}
\gets
\operatorname{ChunkPool}_{k}(\mathcal{S}^{e})$
\Comment{TokenEncoder$\rightarrow$ConceptModule CRC alignment}

\For{$m=1,\ldots,M$}
    \State Split $\mathbf{c}_{m}$ into
    $\mathbf{c}_{m}^{1},\ldots,\mathbf{c}_{m}^{S}$
    \For{$s=1,\ldots,S$}
        \State $n_{m}^{s}
        \gets
        \arg\min_{n\in\{1,\ldots,N\}}
        \left\|
        \mathbf{c}_{m}^{s}-\mathbf{e}_{n}^{s}
        \right\|_{2}^{2}$
        \State $\mathbf{d}_{m}^{s}
        \gets
        \mathbf{e}_{n_{m}^{s}}^{s}$
    \EndFor
\EndFor

\State $\mathcal{L}_{\mathrm{VQ}}
\gets
\frac{1}{MS}
\sum_{m=1}^{M}
\sum_{s=1}^{S}
\left\|
\operatorname{sg}[\mathbf{c}_{m}^{s}]
-\mathbf{d}_{m}^{s}
\right\|_{2}^{2}$

\State $(\mathbf{U},\mathcal{S}^{c})
\gets
\operatorname{ConceptModule}_{\theta_c}^{\mathrm{IRC+CRC}}
\left(
\mathbf{C}_{1:M-1};
\overline{\mathcal{S}}^{e}
\right)$
\Comment{IRC and TokenEncoder$\rightarrow$ConceptModule CRC}

\For{$m=2,\ldots,M$}
    \For{$s=1,\ldots,S$}
        \State $\boldsymbol{\pi}_{m}^{s}
        \gets
        \operatorname{softmax}
        \left(
        \operatorname{PredictionHead}_{s}
        (\mathbf{u}_{m-1})
        \right)$
        \State $\hat{\mathbf{c}}_{m}^{s}
        \gets
        \sum_{n=1}^{N}
        \pi_{m,n}^{s}\mathbf{e}_{n}^{s}$
    \EndFor
    \State $\hat{\mathbf{c}}_{m}
    \gets
    \operatorname{concat}
    \left(
    \hat{\mathbf{c}}_{m}^{1},
    \ldots,
    \hat{\mathbf{c}}_{m}^{S}
    \right)$
\EndFor

\State $\mathcal{L}_{\mathrm{NCP}}
\gets
\frac{1}{(M-1)S}
\sum_{m=2}^{M}
\sum_{s=1}^{S}
\left\|
\hat{\mathbf{c}}_{m}^{s}
-\operatorname{sg}[\mathbf{c}_{m}^{s}]
\right\|_{2}^{2}$

\State $\mathbf{B}
\gets
\operatorname{CausalShiftAndRepeat}
\left(
(\hat{\mathbf{c}}_{2},\ldots,\hat{\mathbf{c}}_{M}),
k,T
\right)$
\Comment{Direct predicted-concept feedback}

\State $\widetilde{\mathbf{H}}
\gets
\mathbf{H}+\mathbf{B}$

\State $\mathcal{S}^{c\rightarrow d}
\gets
\operatorname{CausalShiftAndRepeat}
\left(
\mathcal{S}^{c},k,T
\right)$
\Comment{ConceptModule$\rightarrow$TokenDecoder CRC alignment}

\State $\mathbf{H}^{\mathrm{dec}}
\gets
\operatorname{TokenDecoder}_{\theta_d}^{\mathrm{IRC+CRC}}
\left(
\widetilde{\mathbf{H}};
\mathcal{S}^{e},
\mathcal{S}^{c\rightarrow d}
\right)$
\Comment{IRC and two incoming CRC paths}

\State $\boldsymbol{\ell}
\gets
\operatorname{LMHead}(\mathbf{H}^{\mathrm{dec}})$

\State $\mathcal{L}_{\mathrm{NTP}}
\gets
\operatorname{CE}
\left(
\boldsymbol{\ell}_{1:T-1},
x_{2:T}
\right)$

\State \Return
$\mathcal{L}_{\mathrm{NTP}}
+\alpha\mathcal{L}_{\mathrm{NCP}}
+\beta\mathcal{L}_{\mathrm{VQ}}$
\end{algorithmic}
\end{algorithm}

For free-form tasks, we teacher-force the expert suffix and compute mean target-token negative log-likelihood,
\begin{equation}
  \ell_m(i) = -\frac{1}{|T_i|}\sum_{j\in T_i}
  \log p_m\!\left(y_{ij}\mid x_i,y_{i,<j}\right).
  \label{eq:proxy-100m-item-nll}
\end{equation}

Item-level scores are first averaged within each source, after which source-level means are averaged within each capability leaf. This source-balanced aggregation prevents large benchmarks from dominating solely through their sample counts. Function generation, competition mathematics, academic STEM, and logical reasoning are retained as separate measurements rather than collapsed into a single cross-domain score. For multiple-choice tasks, we use a task-native contrastive statistic: on each HellaSwag item, the character-normalized log-likelihood of the correct continuation is compared with that of the strongest distractor. Both trajectory NLL and choice margin therefore require only batched likelihood evaluation, without autoregressive answer generation or an auxiliary judge model.

\subsection{Recipe-level Results}
\label{sec:proxy-recipe-results}

Downstream results are obtained under a unified evaluation protocol. The Stage-1 model uses the standard pre-training evaluation configuration and is identified by GSM8K = 46.78 and MATH-500 = 14.34. The three Stage-2 variants (V1/V2/V3) share the same inference and sampling configuration, making their relative performance the primary basis for recipe comparison. Stage-1 is used only to provide cross-stage context. In addition to rank agreement, we fit least-squares lines to assess whether differences in Proxy measurements reflect the magnitude of downstream performance differences. Table~\ref{tab:proxy-100m-concept-v123} reports the complete measurements and downstream results used in this analysis.

\begin{table}[htbp]
  \centering
  \small
  \setlength{\tabcolsep}{4pt}
  \renewcommand{\arraystretch}{1.25}
  \caption{Exact inputs to the recipe-screening comparison. Panel (a) reports source-balanced trajectory NLL for the four free-form capability leaves and the task-native correct-versus-best-distractor margin for HellaSwag. Panel (b) reports the corresponding frozen downstream scores. The common prefix in the first header row combines with each Stage suffix to form the exact checkpoint ID. Bold marks the best value among the matched-budget Stage-2 recipes; the Stage-1 column provides cross-stage context and is not part of that selection.}
  \begin{tabular}{>{\centering\arraybackslash}p{4.5cm} *4{>{\centering\arraybackslash}p{2.6cm}}}
    \toprule
    & \multicolumn{4}{c}{\texttt{NCP\_ArchPreview\_dolma3\_8.9B\_}} \\
    \cmidrule(lr){2-5}
    Metric
      & Stage1\_V1
      & Stage2\_V1
      & Stage2\_V2
      & Stage2\_V3 \\
    \midrule
    \multicolumn{5}{@{}l}{\textit{(a) Expanded 100M Proxy measurements}} \\
    Function-generation NLL $\downarrow$
      & \ProxyCodeStageOne
      & \textbf{\ProxyCodeStageTwoVOne}
      & \ProxyCodeStageTwoVTwo
      & \ProxyCodeStageTwoVThree \\
    Competition-math NLL $\downarrow$
      & \ProxyCompetitionMathStageOne
      & \textbf{\ProxyCompetitionMathStageTwoVOne}
      & \ProxyCompetitionMathStageTwoVTwo
      & \ProxyCompetitionMathStageTwoVThree \\
    Academic-STEM NLL $\downarrow$
      & \ProxyAcademicSTEMStageOne
      & \textbf{\ProxyAcademicSTEMStageTwoVOne}
      & \ProxyAcademicSTEMStageTwoVTwo
      & \ProxyAcademicSTEMStageTwoVThree \\
    Logical-reasoning NLL $\downarrow$
      & \ProxyLogicalReasoningStageOne
      & \textbf{\ProxyLogicalReasoningStageTwoVOne}
      & \ProxyLogicalReasoningStageTwoVTwo
      & \ProxyLogicalReasoningStageTwoVThree \\
    HellaSwag margin $\uparrow$
      & \ProxyHellaMarginStageOne
      & \ProxyHellaMarginStageTwoVOne
      & \ProxyHellaMarginStageTwoVTwo
      & \textbf{\ProxyHellaMarginStageTwoVThree} \\
    \addlinespace[2pt]
    \multicolumn{5}{@{}l}{\textit{(b) Frozen downstream performance (\%)}} \\
    HumanEval $\uparrow$
      & \DownstreamHumanEvalStageOne
      & \textbf{\DownstreamHumanEvalStageTwoVOne}
      & \DownstreamHumanEvalStageTwoVTwo
      & \DownstreamHumanEvalStageTwoVThree \\
    MBPP $\uparrow$
      & \DownstreamMBPPStageOne
      & \textbf{\DownstreamMBPPStageTwoVOne}
      & \DownstreamMBPPStageTwoVTwo
      & \DownstreamMBPPStageTwoVThree \\
    MATH-500 $\uparrow$
      & \DownstreamMATHFiveHundredStageOne
      & \textbf{\DownstreamMATHFiveHundredStageTwoVOne}
      & \DownstreamMATHFiveHundredStageTwoVTwo
      & \DownstreamMATHFiveHundredStageTwoVThree \\
    Minerva $\uparrow$
      & \DownstreamMinervaStageOne
      & \textbf{\DownstreamMinervaStageTwoVOne}
      & \DownstreamMinervaStageTwoVTwo
      & \DownstreamMinervaStageTwoVThree \\
    MMLU-STEM $\uparrow$
      & \DownstreamMMLUSTEMStageOne
      & \textbf{\DownstreamMMLUSTEMStageTwoVOne}
      & \DownstreamMMLUSTEMStageTwoVTwo
      & \DownstreamMMLUSTEMStageTwoVThree \\
    BBH $\uparrow$
      & \DownstreamBBHStageOne
      & \textbf{\DownstreamBBHStageTwoVOne}
      & \DownstreamBBHStageTwoVTwo
      & \DownstreamBBHStageTwoVThree \\
    HellaSwag $\uparrow$
      & \DownstreamHellaSwagStageOne
      & \DownstreamHellaSwagStageTwoVOne
      & \textbf{\DownstreamHellaSwagStageTwoVTwo}
      & \DownstreamHellaSwagStageTwoVThree \\
    \bottomrule
  \end{tabular}
  \label{tab:proxy-100m-concept-v123}
\end{table}

\begin{figure*}[htbp]
  \centering
  \includegraphics[width=\textwidth]{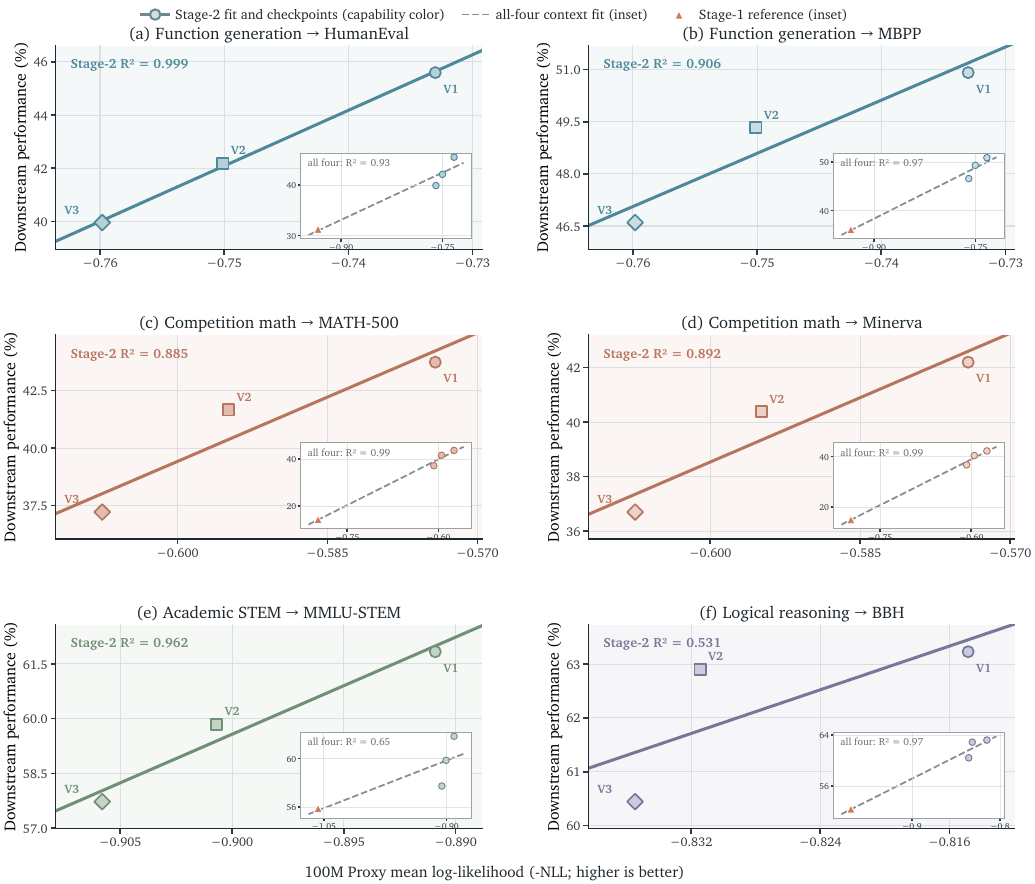}
  \caption{Relationship between the expanded 100M Proxy and downstream benchmark performance. The main panels show the three same-protocol Stage-2 recipes. The horizontal axis is source-balanced mean log-likelihood ($-\mathrm{NLL}$, higher is better), and the vertical axis is downstream performance in percent. Insets add the Stage-1 reference and the corresponding four-point fit for context. Each capability--benchmark mapping is reported separately; the lines are descriptive fits rather than calibrated predictors.}
  \label{fig:proxy-100m-downstream-fit}
\end{figure*}

Across all four free-form capability leaves, the Stage-2 variants exhibit a consistent ordering: V1 achieves the lowest NLL, followed by V2 and V3. The corresponding downstream benchmarks show the same ordering: HumanEval and MBPP for code, MATH-500 and Minerva for mathematics, MMLU-STEM for academic STEM, and BBH for logical reasoning. Thus, each capability--benchmark mapping is concordant across all three pairwise Stage-2 comparisons. Relative to V3, V1 reduces NLL by \ProxyCodeStageTwoVOneVersusVThreeReduction\ for function generation, \ProxyCompetitionMathStageTwoVOneVersusVThreeReduction\ for competition mathematics, \ProxyAcademicSTEMStageTwoVOneVersusVThreeReduction\ for academic STEM, and \ProxyLogicalReasoningStageTwoVOneVersusVThreeReduction\ for logical reasoning. These reductions correspond to downstream improvements of \DownstreamHumanEvalVOneMinusVThree\ points on HumanEval, \DownstreamMBPPVOneMinusVThree\ on MBPP, \DownstreamMATHFiveHundredVOneMinusVThree\ on MATH-500, \DownstreamMinervaVOneMinusVThree\ on Minerva, \DownstreamMMLUSTEMVOneMinusVThree\ on MMLU-STEM, and \DownstreamBBHVOneMinusVThree\ on BBH. The same four Proxy orderings are preserved when the analysis is repeated on the trusted subset.

\subsection{Interpretation and Scope}
\label{sec:proxy-interpretation}

The fitted relationships in Figure~\ref{fig:proxy-100m-downstream-fit} complement the rank comparison by characterizing the relative magnitudes of the observed differences. The Stage-2 $R^2$ values are \ProxyFitStageTwoRSquaredHumanEval\ for HumanEval, \ProxyFitStageTwoRSquaredMBPP\ for MBPP, \ProxyFitStageTwoRSquaredMATHFiveHundred\ for MATH-500, \ProxyFitStageTwoRSquaredMinerva\ for Minerva, and \ProxyFitStageTwoRSquaredMMLUSTEM\ for MMLU-STEM. BBH exhibits a lower $R^2$ of \ProxyFitStageTwoRSquaredBBH: its recipe ordering remains concordant, although the Proxy differences are not proportional to the downstream score differences. Taken together, the matched-budget comparisons identify V1 as the strongest of the three Stage-2 recipes across the evaluated free-form capabilities. This conclusion applies at the level of complete data recipes; it does not attribute the observed improvements to any individual source within a mixture.

HellaSwag further illustrates the importance of matching the scoring statistic to the task format. The task-native choice margin increases from \ProxyHellaMarginStageTwoVOne\ for V1 to \ProxyHellaMarginStageTwoVTwo\ for V2 and \ProxyHellaMarginStageTwoVThree\ for V3. The downstream scores place V2 and V3 in the leading tier at \DownstreamHellaSwagStageTwoVTwo\ and \DownstreamHellaSwagStageTwoVThree, respectively, compared with \DownstreamHellaSwagStageTwoVOne\ for V1. Because V2 and V3 differ by only \CoreHellaVTwoMinusVThree\ percentage points downstream, we treat them as effectively tied for recipe screening rather than infer a preference from this negligible gap. At this decision granularity, the choice margin distinguishes the leading V2/V3 tier from V1 and provides a more appropriate summary for multiple-choice evaluation.

In practice, a new mid-training recipe can be screened with a single frozen forward pass by comparing its capability-level NLL values and task-native margins with those of existing recipes. Retaining these measurements as a capability vector supports targeted recipe selection---for example, emphasizing code and mathematics---without imposing an arbitrary global aggregation across domains. The present evidence covers three Stage-2 recipes and one cross-stage reference. It therefore supports efficient recipe screening within the evaluated setting, but does not establish calibrated prediction for unseen model families or causal attribution to individual data components.

\section{Scaling Experiments}\label{app:scaling-experiments}

This section describes how we conduct the scaling-ladder experiments. We first
select several FLOPs budgets and design multiple allocations between model size and training tokens for each budget. Although the formula
$C\approx 6N_{\mathrm{param}}D$, where $N_{\mathrm{param}}$ denotes the number
of non-embedding parameters, has been widely adopted as an approximation of
training computation, it can be inaccurate for architectures with non-uniform layer activation. In \texttt{NCP-ArchPreview}, the Concept Module operates only once per chunk. We therefore
compute the total training budget as
\[
    C = F_{\mathrm{tok}}D,
\]
where $F_{\mathrm{tok}}$ denotes the analytical training FLOPs per token and
$D$ denotes the number of training tokens. The FLOPs estimates exclude
embedding parameters.

\begin{figure}[tbp]
        \centering
        \includegraphics[width=0.6\linewidth]{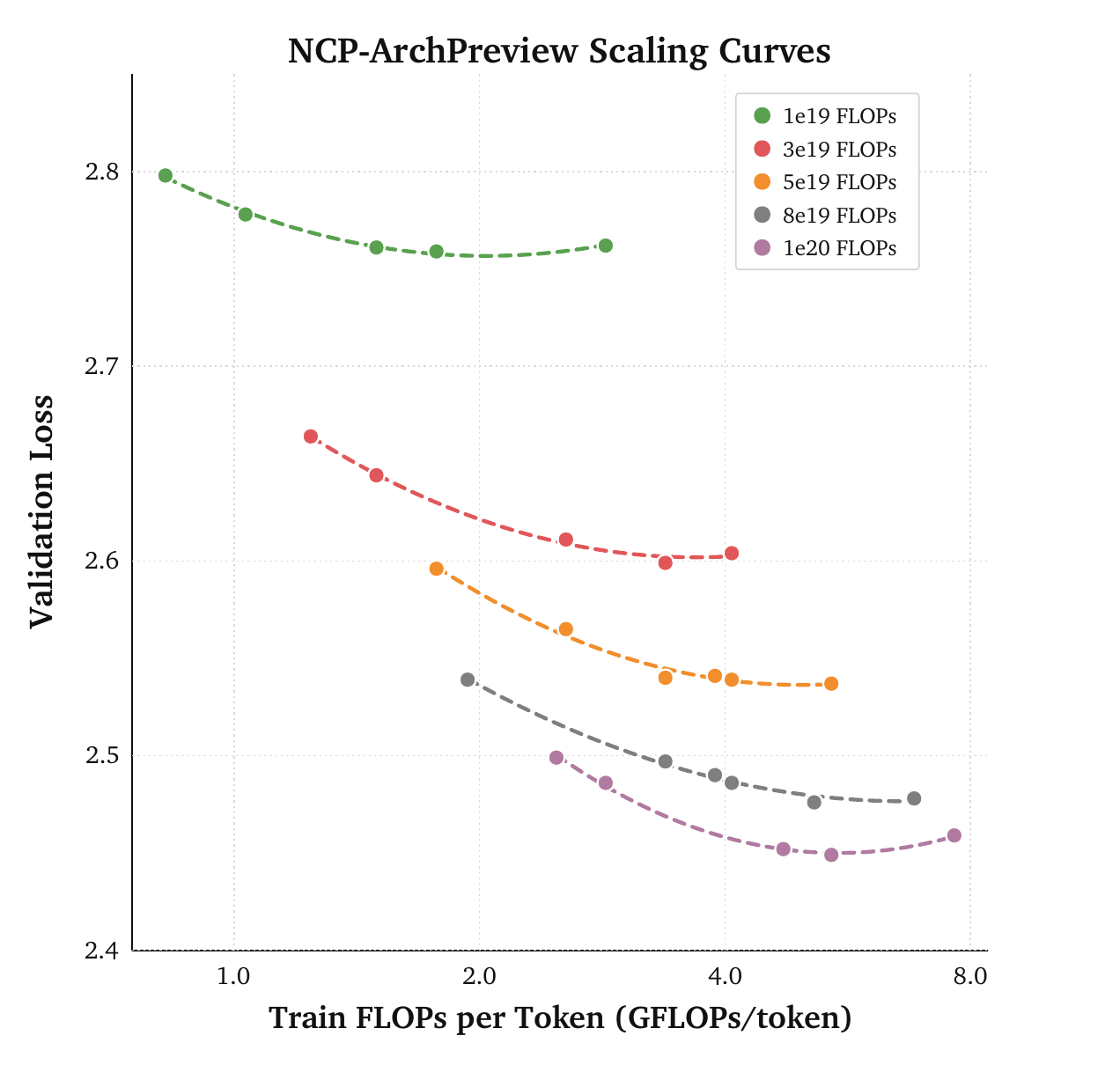}
        \caption{\texttt{NCP-ArchPreview} IsoFLOP Curves. Markers denote experimental results, while dashed curves show the fitted trends through the measured points.}
        \label{fig:appendix-conceptlm-scaling-curves}
\end{figure}

Previous study~\citep{deepseekai2024deepseekllm} finds that different experiments under different model/data allocations tend to have similar optimal hyperparameters at a fixed FLOPs
budget. 
Motivated by this observation, we first choose a
representative model at each FLOPs budget and perform search over learning
rate and batch size on these models. After determining the optimal learning rate, we search
for the best learning rate of the other models at the same FLOPs budget in a
neighborhood around it. Each point in Figure~\ref{fig:appendix-conceptlm-scaling-curves} reports
the result obtained using the best searched hyperparameters for the
corresponding model and FLOPs budget. 
Table~\ref{tab:scaling-experiment-settings} lists the experimental settings for every point shown in the figure.

\begin{table}[tbp]
    \centering
\caption{\texttt{NCP-ArchPreview} scaling-ladder experiment settings and validation losses. Here,
$H$ is the hidden width, and $L_e/L_c/L_d$ are the
encoder/Concept Module/decoder depths, $F_{\mathrm{tok}}$ is the analytical training FLOPs
per token, and $D$ is the training-token count. The reported loss is the lowest validation loss among the
searched learning rates for each point.}\label{tab:scaling-experiment-settings}
    \resizebox{0.8\linewidth}{!}{%
    \begin{tabular}{@{}c c c c c c c c@{}}
    \toprule
    $C$ (FLOPs) & Pt. & Heads & $H$ & $L_e/L_c/L_d$ & $F_{\mathrm{tok}}$ (GF/token) & $D$ (B) & Val. loss\\
    \midrule
    1e19 & 1 & 7 & 896 & 4/4/4 & 0.824 & 12.140 & 2.798 \\
    1e19 & 2 & 8 & 1024 & 4/4/4 & 1.033 & 9.676 & 2.778 \\
    1e19 & 3 & 10 & 1280 & 4/4/4 & 1.495 & 6.688 & 2.761 \\
    1e19 & 4 & 11 & 1408 & 4/4/4 & 1.771 & 5.645 & 2.759 \\
    1e19 & 5 & 10 & 1280 & 8/4/8 & 2.856 & 3.502 & 2.762 \\
    \midrule
    3e19 & 1 & 9 & 1152 & 4/4/4 & 1.242 & 24.150 & 2.664 \\
    3e19 & 2 & 10 & 1280 & 4/4/4 & 1.495 & 20.063 & 2.644 \\
    3e19 & 3 & 11 & 1408 & 6/3/6 & 2.553 & 11.752 & 2.611 \\
    3e19 & 4 & 11 & 1408 & 8/4/8 & 3.380 & 8.874 & 2.599 \\
    3e19 & 5 & 12 & 1536 & 8/8/8 & 4.077 & 7.358 & 2.604 \\
    \midrule
    5e19 & 1 & 11 & 1408 & 4/4/4 & 1.771 & 28.227 & 2.596 \\
    5e19 & 2 & 11 & 1408 & 6/3/6 & 2.553 & 19.587 & 2.565 \\
    5e19 & 3 & 11 & 1408 & 8/4/8 & 3.380 & 14.791 & 2.540 \\
    5e19 & 4 & 12 & 1536 & 8/4/8 & 3.888 & 12.860 & 2.541 \\
    5e19 & 5 & 12 & 1536 & 8/8/8 & 4.077 & 12.264 & 2.539 \\
    5e19 & 6 & 14 & 1792 & 8/8/8 & 5.401 & 9.258 & 2.537 \\
    \midrule
    8e19 & 1 & 11 & 1408 & 4/8/4 & 1.934 & 41.375 & 2.539 \\
    8e19 & 2 & 11 & 1408 & 8/4/8 & 3.380 & 23.665 & 2.497 \\
    8e19 & 3 & 12 & 1536 & 8/4/8 & 3.888 & 20.574 & 2.490 \\
    8e19 & 4 & 12 & 1536 & 8/8/8 & 4.077 & 19.622 & 2.486 \\
    8e19 & 5 & 14 & 1792 & 8/4/8 & 5.145 & 15.550 & 2.476 \\
    8e19 & 6 & 16 & 2048 & 8/8/8 & 6.823 & 11.724 & 2.478 \\
    \midrule
    1e20 & 1 & 9 & 1152 & 8/8/8 & 2.485 & 40.249 & 2.499 \\
    1e20 & 2 & 10 & 1280 & 8/4/8 & 2.856 & 35.015 & 2.486 \\
    1e20 & 3 & 13 & 1664 & 8/8/8 & 4.716 & 21.205 & 2.452 \\
    1e20 & 4 & 14 & 1792 & 8/8/8 & 5.401 & 18.516 & 2.449 \\
    1e20 & 5 & 17 & 2176 & 8/8/8 & 7.641 & 13.087 & 2.459 \\
    \bottomrule
    \end{tabular}%
    }
\end{table}

\section{Evaluation Details}
\label{app:evaluation-details}

\paragraph{Benchmark coverage:}
The main comparison in Table~\ref{tab:core88-results} separates
higher-is-better task scores from lower-is-better likelihood scores.
The higher-is-better portion reports the following benchmark groups:

\begin{itemize}
    \setlength{\itemsep}{2pt}

    \item \textbf{MMLU:}
    MMLU-Humanities, MMLU-Social Sciences, MMLU-STEM, MMLU-Other,
    and the aggregate MMLU score~\citep{hendrycks2021mmlu}, together
    with MMLU-Pro~\citep{wang2024mmlupro}.

    \item \textbf{Mathematical reasoning:}
    GSM8K~\citep{cobbe2021gsm8k},
    GSM-Symbolic~\citep{mirzadeh2025gsmsymbolic},
    Minerva Math~\citep{hendrycks2021math,lewkowycz2022minerva}, and
    MATH-500~\citep{hendrycks2021math,lightman2023verify}.

    \item \textbf{Code generation:}
    BigCodeBench~\citep{zhuo2025bigcodebench},
    HumanEval~\citep{chen2021humaneval},
    DS-1000~\citep{lai2023ds1000},
    MBPP~\citep{austin2021mbpp}, and the HumanEval and MBPP variants
    of MultiPL-E~\citep{cassano2022multiple}.

    \item \textbf{Multiple-choice STEM reasoning:}
    ARC-Easy and ARC-Challenge~\citep{clark2018arc}.

    \item \textbf{Multiple-choice non-STEM reasoning:}
    PIQA~\citep{bisk2020piqa},
    CommonsenseQA~\citep{talmor2019commonsenseqa}, and
    SocialIQA~\citep{sap2019socialiqa}.

    \item \textbf{General question answering:}
    HellaSwag~\citep{zellers2019hellaswag},
    WinoGrande~\citep{sakaguchi2020winogrande},
    LAMBADA-Standard and LAMBADA-OpenAI~\citep{paperno2016lambada},
    MedMCQA~\citep{pal2022medmcqa}, and
    MedQA~\citep{jin2021medqa}.
\end{itemize}

The lower-is-better likelihood portion reports bits per byte (BPB) on
HumanEval Gold~\citep{chen2021humaneval},
MBPP Gold~\citep{austin2021mbpp},
MT-MBPP Gold~\citep{cassano2022multiple},
CoQA~\citep{reddy2019coqa},
DROP~\citep{dua2019drop},
GSM8K Gold~\citep{cobbe2021gsm8k},
Jeopardy~\citep{soldni2024jeopardy},
LAMBADA-OpenAI~\citep{paperno2016lambada},
Natural Questions~\citep{kwiatkowski2019naturalquestions}, and
SQuAD~\citep{rajpurkar2016squad}.
The ``Gold'' suffix denotes evaluation of the likelihood assigned to a
reference completion rather than free-form generation.

The group labels above are used only to organize the reported results
and do not imply that all benchmarks within a group share the same
evaluation format. In particular, the GenQA group contains both
likelihood-ranked choice tasks and zero-shot language-modeling tasks.

We use a global random seed of 42 for few-shot example selection and
generation. Prompt templates, demonstrations, answer choices, stop
sequences, evaluation splits, and scoring implementations are held fixed
across all evaluated models.

\paragraph{Multiple-choice and language-modeling accuracy evaluation.}
The four MMLU domain scores and the aggregate MMLU score are obtained
using 5-shot multiple-choice evaluation on the test split. Each domain
score is macro-averaged equally across the MMLU subjects assigned to that
domain, while the aggregate MMLU score is macro-averaged equally across
all 57 subjects. MMLU-Pro is evaluated using its 5-shot multiple-choice
formulation.

ARC-Easy, ARC-Challenge, CommonsenseQA, PIQA, and SocialIQA are evaluated
using 5-shot multiple-choice prompts and raw choice accuracy. HellaSwag
uses a 5-shot ranked-choice formulation with answer likelihood normalized
by character length, whereas WinoGrande uses a 5-shot ranked-choice
formulation with raw accuracy. MedMCQA and MedQA use 5-shot ranked-choice
evaluation with answer likelihood normalized by token length.
LAMBADA-Standard and LAMBADA-OpenAI are evaluated zero-shot using greedy
next-token completion accuracy in the higher-is-better portion of the
table.

\paragraph{Gold-completion and open-ended likelihood evaluation.}
The likelihood evaluation uses three demonstrations for HumanEval Gold,
MBPP Gold, and every MT-MBPP language; zero demonstrations for CoQA and
LAMBADA-OpenAI; five demonstrations for DROP, Jeopardy, Natural
Questions, and SQuAD; and eight demonstrations for GSM8K Gold. Each task
is summarized by the mean negative base-2 log-likelihood of its gold
continuations normalized by their UTF-8 byte lengths, measured in bits
per byte. The three code likelihood benchmarks use the corrected BPB
implementation provided by the evaluation framework. Lower BPB indicates
better performance. These results and their AVG are kept separate from
the higher-is-better task scores and Overall AVG.

\paragraph{Generative evaluation.}
Table~\ref{tab:generation-configurations} lists the prompting and
sampling configurations for benchmarks that require free-form
generation. The number of samples denotes generated completions per
problem and is independent of the inference batch size.

\begin{table*}[t]
\centering
\small
\setlength{\tabcolsep}{4pt}
\renewcommand{\arraystretch}{1.10}
\caption{Prompting and sampling configurations for the generative
benchmarks reported in Table~\ref{tab:core88-results}. The number of
samples denotes generated completions per problem.}
\label{tab:generation-configurations}
\begin{tabular}{@{}l c c c@{}}
\toprule
Benchmark & Shots & Samples & Metric \\
\midrule

\multicolumn{4}{@{}l}{\textit{Mathematical reasoning}} \\
GSM8K                 & 8 & 1  & Pass@1 \\
GSM-Symbolic          & 8 & 8  & Pass@1 \\
Minerva Math          & 4 & 1  & Exact match \\
MATH-500              & 4 & 32 & Pass@1 \\

\midrule
\multicolumn{4}{@{}l}{\textit{Code generation}} \\
BigCodeBench          & 3 & 5  & Execution pass@1 \\
HumanEval             & 3 & 32 & Execution pass@1 \\
DS-1000               & 3 & 5  & Execution pass@1 \\
MBPP                   & 3 & 32 & Execution pass@1 \\
MultiPL-E HumanEval   & 0 & 32 & Execution pass@1 \\
MultiPL-E MBPP        & 0 & 32 & Execution pass@1 \\

\bottomrule
\end{tabular}
\end{table*}

\paragraph{Scoring and aggregation.}
Mathematical answers are normalized using benchmark-specific answer
extractors. GSM8K and GSM-Symbolic use last-number extraction, while
Minerva Math and MATH-500 use their corresponding normalized
mathematical equivalence rules. The GSM-Symbolic score is macro-averaged
equally across its three subsets, and the Minerva Math score is
macro-averaged equally across its seven subject categories.

Code completions are executed in isolated task-specific environments
using the corresponding official test cases and dependencies. The
MultiPL-E HumanEval and MultiPL-E MBPP scores are each macro-averaged
equally across the six reported programming languages: C++, C\#, Go,
Java, JavaScript, and PHP.

The higher-is-better results in Table~\ref{tab:core88-results} are
reported as percentage points. Within that portion of the table, each
section AVG is the unweighted mean of the following constituent results:

\begin{itemize}
    \setlength{\itemsep}{2pt}

    \item \textbf{MMLU AVG:}
    MMLU-Humanities, MMLU-Social Sciences, MMLU-STEM, MMLU-Other,
    and MMLU-Pro. The aggregate MMLU row is excluded because it
    summarizes the same 57 MMLU subjects.

    \item \textbf{MATH AVG:}
    GSM8K, GSM-Symbolic, Minerva Math, and MATH-500.

    \item \textbf{Code AVG:}
    BigCodeBench, HumanEval, DS-1000, MBPP,
    MultiPL-E HumanEval, and MultiPL-E MBPP.

    \item \textbf{MC-STEM AVG:}
    ARC-Easy and ARC-Challenge.

    \item \textbf{MC-Non-STEM AVG:}
    PIQA, CommonsenseQA, and SocialIQA.

    \item \textbf{GenQA AVG:}
    HellaSwag, WinoGrande, LAMBADA-Standard, LAMBADA-OpenAI,
    MedMCQA, and MedQA.
\end{itemize}

The higher-is-better Overall AVG is the unweighted mean of the 26
constituent benchmark results listed above, rather than the mean of the
section AVGs. It excludes the aggregate MMLU row, which is shown only as
a summary of the 57 MMLU subjects, and excludes all intermediate AVG
rows.

For MT-MBPP Gold, BPB is first macro-averaged equally across its 17
language subsets. The lower-is-better AVG is then the unweighted mean of
the ten benchmark-level BPB results listed in the likelihood portion of
the table. Consequently, MT-MBPP Gold receives the same weight as each
of the other nine benchmarks, rather than 17 times their weight. The BPB
AVG is computed independently and is excluded from every domain AVG and
from the higher-is-better Overall AVG.

Every displayed delta is computed as the NCP score minus the
corresponding Vanilla score. Therefore, a positive delta denotes an
improvement for the higher-is-better task scores, whereas a negative
delta denotes an improvement for the lower-is-better BPB scores.

\paragraph{Domain-adaptation evaluation.}
For the VQ training study in Section~\ref{sec:vq-training}, the
code-adapted model is evaluated on HumanEval~\citep{chen2021humaneval}
and MBPP~\citep{austin2021mbpp} using the augmented HumanEval+ and MBPP+
test suites provided by EvalPlus~\citep{liu2023evalplus}. The
mathematics-adapted model is evaluated on GSM8K~\citep{cobbe2021gsm8k}
and MATH-500~\citep{hendrycks2021math,lightman2023verify}. The
knowledge-adapted model is evaluated on
TriviaQA~\citep{joshi2017triviaqa} using its standard answer-matching
protocol. General-capability retention is measured using
MMLU~\citep{hendrycks2021mmlu}, ARC-Easy~\citep{clark2018arc},
HellaSwag~\citep{zellers2019hellaswag}, and Natural
Questions~\citep{kwiatkowski2019naturalquestions}. These auxiliary
domain-adaptation results are not included in
Table~\ref{tab:core88-results} or either of its aggregate scores.

\paragraph{Detailed results on the stage-1 checkpoint trajectory:}
Table~\ref{tab:stage1-checkpoint-trajectory} shows detailed results on the stage-1 checkpoint trajectory.

\begin{table*}[p]
    \centering
    \caption{
        \texttt{NCP-ArchPreview} stage-1 downstream performance across training
        checkpoints. The reported benchmarks follow the same grouping
        as Table~\ref{tab:core88-results}. The likelihood block reports BPB (lower is better) and is aggregated separately from the higher-is-better Overall AVG.
    }
    \label{tab:stage1-checkpoint-trajectory}

    \setlength{\tabcolsep}{2pt}

    \resizebox{\textwidth}{!}{
    \begin{tabular}{@{}l*{14}{r}@{}}
        \toprule
        Dataset
        & 100K & 200K & 300K & 400K & 500K & 600K & 700K
        & 800K & 900K & 1.0M & 1.1M & 1.2M & 1.3M & Final \\
        \midrule

        \textbf{MMLU} \\
        MMLU-Humanities
        & 54.38 & 58.12 & 62.77 & 63.02 & 63.57 & 65.10 & 64.57
        & 65.14 & 65.69 & 66.16 & 66.64 & 67.74 & 68.26 & 68.28 \\
        MMLU-Social Sciences
        & 60.38 & 64.61 & 66.97 & 67.09 & 69.65 & 69.92 & 70.54
        & 71.57 & 72.05 & 71.73 & 72.44 & 73.62 & 73.41 & 73.97 \\
        MMLU-STEM
        & 46.06 & 49.17 & 50.32 & 50.89 & 53.96 & 51.47 & 53.62
        & 53.86 & 54.48 & 54.12 & 54.43 & 54.05 & 55.45 & 55.53 \\
        MMLU-Other
        & 54.65 & 58.13 & 61.53 & 61.78 & 63.64 & 62.89 & 64.23
        & 65.42 & 65.07 & 65.46 & 66.04 & 66.93 & 66.10 & 65.63 \\
        MMLU
        & 53.08 & 56.66 & 59.42 & 59.74 & 61.83 & 61.27 & 62.29
        & 63.00 & 63.34 & 63.36 & 63.86 & 64.46 & 64.77 & 64.80 \\
        MMLU-Pro
        & 15.59 & 16.89 & 17.15 & 18.46 & 18.65 & 18.73 & 19.27
        & 19.30 & 19.66 & 20.45 & 20.15 & 20.17 & 20.25 & 20.26 \\
        \midrule[\heavyrulewidth]

        \textbf{MATH} \\
        GSM8K
        & 23.43 & 31.16 & 32.75 & 34.87 & 39.04 & 40.49 & 40.26
        & 41.77 & 42.15 & 43.37 & 42.38 & 44.58 & 45.34 & 45.26 \\
        GSM-Symbolic
        & 8.55 & 13.14 & 14.37 & 15.88 & 17.12 & 19.13 & 19.32
        & 20.12 & 20.81 & 21.86 & 20.37 & 22.05 & 23.05 & 22.80 \\
        Minerva
        & 7.49 & 8.71 & 10.45 & 10.51 & 11.70 & 12.26 & 11.43
        & 12.46 & 13.76 & 14.27 & 14.46 & 14.83 & 14.62 & 15.63 \\
        MATH-500
        & 6.98 & 7.93 & 9.77 & 9.90 & 10.42 & 10.36 & 11.16
        & 11.99 & 12.67 & 12.65 & 13.81 & 13.73 & 13.66 & 14.48 \\
        \midrule[\heavyrulewidth]

        \textbf{Code} \\
        BigCodeBench
        & 10.56 & 14.70 & 17.61 & 16.26 & 19.04 & 19.51 & 19.14
        & 20.46 & 21.42 & 21.75 & 20.93 & 21.93 & 22.54 & 22.79 \\
        HumanEval
        & 20.05 & 23.30 & 27.02 & 25.06 & 25.93 & 26.14 & 28.87
        & 26.33 & 29.55 & 30.30 & 30.53 & 30.30 & 29.40 & 31.38 \\
        DS-1000
        & 8.84 & 12.74 & 14.44 & 13.60 & 16.12 & 16.96 & 16.90
        & 17.00 & 18.70 & 19.70 & 18.80 & 19.86 & 20.42 & 20.40 \\
        MBPP
        & 27.71 & 27.46 & 31.15 & 31.86 & 30.43 & 31.66 & 33.26
        & 34.59 & 35.29 & 35.04 & 34.99 & 36.64 & 36.09 & 35.91 \\
        MultiPL-E HumanEval
        & 15.17 & 16.81 & 18.26 & 18.44 & 18.16 & 19.18 & 20.33
        & 22.38 & 22.17 & 21.18 & 21.26 & 22.07 & 21.36 & 22.27 \\
        MultiPL-E MBPP
        & 27.64 & 27.15 & 30.34 & 29.54 & 29.86 & 32.20 & 32.04
        & 34.65 & 34.00 & 34.18 & 33.36 & 34.64 & 34.86 & 33.96 \\
        \midrule[\heavyrulewidth]

        \textbf{MC-STEM} \\
        ARC-E
        & 82.24 & 86.24 & 88.38 & 90.19 & 89.10 & 90.40 & 91.84
        & 91.84 & 92.05 & 91.88 & 92.05 & 92.26 & 92.97 & 92.30 \\
        ARC-C
        & 65.61 & 69.97 & 72.78 & 74.74 & 75.43 & 76.11 & 77.30
        & 78.16 & 79.27 & 78.67 & 78.92 & 81.31 & 80.46 & 81.57 \\
        \midrule[\heavyrulewidth]

        \textbf{MC-Non-STEM} \\
        PiQA
        & 67.08 & 72.20 & 73.61 & 75.95 & 76.12 & 77.58 & 77.58
        & 79.54 & 80.20 & 79.98 & 79.92 & 80.85 & 80.96 & 80.85 \\
        CommonsenseQA
        & 62.82 & 68.80 & 70.76 & 70.52 & 71.99 & 71.42 & 72.73
        & 73.55 & 73.71 & 73.55 & 73.14 & 73.14 & 74.69 & 74.04 \\
        SocialIQA
        & 59.42 & 63.56 & 65.30 & 66.79 & 67.30 & 67.91 & 69.09
        & 69.70 & 68.12 & 68.78 & 69.09 & 69.50 & 69.45 & 69.24 \\
        \midrule[\heavyrulewidth]

        \textbf{GenQA} \\
        HellaSwag
        & 61.60 & 62.70 & 63.60 & 64.75 & 63.55 & 64.10 & 65.20
        & 66.05 & 65.35 & 66.05 & 66.05 & 65.70 & 66.00 & 66.05 \\
        WinoGrande
        & 50.59 & 49.72 & 51.14 & 49.25 & 50.59 & 48.54 & 49.88
        & 49.57 & 49.41 & 50.28 & 48.78 & 49.25 & 50.51 & 51.22 \\
        LAMBADA-Standard
        & 62.45 & 63.34 & 64.60 & 65.90 & 66.00 & 66.02 & 67.86
        & 66.89 & 65.90 & 65.03 & 67.01 & 65.85 & 66.54 & 66.89 \\
        LAMBADA-OpenAI
        & 67.46 & 68.83 & 68.29 & 70.81 & 70.83 & 70.79 & 71.61
        & 71.67 & 70.42 & 71.47 & 71.80 & 72.00 & 71.51 & 72.21 \\
        MedMCQA
        & 30.98 & 31.81 & 31.80 & 31.64 & 32.08 & 32.14 & 32.32
        & 32.69 & 32.88 & 32.91 & 32.69 & 33.15 & 33.16 & 33.21 \\
        MedQA
        & 32.99 & 34.72 & 35.74 & 36.53 & 36.92 & 36.53 & 37.55
        & 38.88 & 39.28 & 38.57 & 37.94 & 38.49 & 39.67 & 38.96 \\
        \midrule[\heavyrulewidth]
        
        \textbf{AVG}
        & \textbf{39.64} & \textbf{42.38} & \textbf{44.27}
        & \textbf{44.78} & \textbf{45.66} & \textbf{46.06}
        & \textbf{46.84} & \textbf{47.52} & \textbf{47.85}
        & \textbf{48.05} & \textbf{48.00} & \textbf{48.64}
        & \textbf{48.87} & \textbf{49.04} \\
        \midrule[\heavyrulewidth]
        
        \textbf{Likelihood (BPB; $\downarrow$)} \\
        HumanEval Gold
        & 0.374 & 0.382 & 0.376 & 0.358 & 0.370 & 0.381 & 0.375
        & 0.361 & 0.366 & 0.376 & 0.364 & 0.359 & 0.363 & 0.364 \\
        MBPP Gold
        & 0.515 & 0.513 & 0.505 & 0.500 & 0.495 & 0.493 & 0.495
        & 0.487 & 0.489 & 0.485 & 0.473 & 0.473 & 0.479 & 0.473 \\
        MT-MBPP Gold
        & 0.472 & 0.461 & 0.445 & 0.446 & 0.448 & 0.447 & 0.457
        & 0.453 & 0.449 & 0.441 & 0.429 & 0.440 & 0.434 & 0.434 \\
        CoQA
        & 0.403 & 0.404 & 0.377 & 0.384 & 0.382 & 0.377 & 0.374
        & 0.373 & 0.379 & 0.366 & 0.372 & 0.357 & 0.367 & 0.370 \\
        DROP
        & 4.922 & 4.831 & 4.985 & 4.512 & 4.592 & 4.479 & 4.424
        & 4.365 & 4.464 & 4.444 & 4.424 & 4.304 & 4.252 & 4.397 \\
        GSM8K Gold
        & 0.431 & 0.419 & 0.418 & 0.402 & 0.406 & 0.405 & 0.406
        & 0.402 & 0.404 & 0.403 & 0.398 & 0.394 & 0.404 & 0.400 \\
        Jeopardy
        & 0.429 & 0.395 & 0.383 & 0.391 & 0.393 & 0.388 & 0.374
        & 0.372 & 0.380 & 0.365 & 0.360 & 0.355 & 0.350 & 0.347 \\
        LAMBADA-OpenAI
        & 0.332 & 0.323 & 1.061 & 0.304 & 0.301 & 0.300 & 0.295
        & 0.293 & 0.297 & 0.296 & 0.291 & 0.289 & 0.291 & 0.288 \\
        Natural Questions
        & 0.993 & 0.969 & 0.957 & 0.939 & 0.942 & 0.932 & 0.917
        & 0.930 & 0.923 & 0.914 & 0.915 & 0.892 & 0.895 & 0.884 \\
        SQuAD
        & 0.162 & 0.159 & 0.158 & 0.154 & 0.154 & 0.135 & 0.146
        & 0.143 & 0.138 & 0.148 & 0.151 & 0.149 & 0.145 & 0.151 \\
        \midrule[\heavyrulewidth]
        \textbf{AVG ($\downarrow$)}
        & \textbf{0.903} & \textbf{0.886} & \textbf{0.967}
        & \textbf{0.839} & \textbf{0.848} & \textbf{0.834}
        & \textbf{0.826} & \textbf{0.818} & \textbf{0.829}
        & \textbf{0.824} & \textbf{0.818} & \textbf{0.801}
        & \textbf{0.798} & \textbf{0.811} \\
        
        \bottomrule
    \end{tabular}
    }
\end{table*}

\end{document}